\documentclass[journal]{IEEEtran}

\usepackage[pagebackref=true,colorlinks,bookmarks=false]{hyperref}

\ifCLASSINFOpdf
\else
	\usepackage{epsfig}
	\usepackage{graphicx}
  \usepackage{subcaption}
	\usepackage[justification=centering]{caption}
	\usepackage{array}
	\newcolumntype{P}[1]{>{\centering\arraybackslash}p{#1}}
\fi

\usepackage{amsmath,amssymb}
\usepackage{algorithm}
\usepackage{algpseudocode}
\usepackage{multirow}

\begin{document}

\title{Efficiently Computing Piecewise Flat Embeddings for Data Clustering and Image Segmentation}

\author{Renee~T.~Meinhold, Tyler~L.~Hayes, and Nathan~D.~Cahill\\
Image Computing \& Analysis Laboratory, School of Mathematical Sciences\\
Rochester Institute of Technology, Rochester, NY, USA\\
\{rtm9271, tlh6792, ndcsma\}@rit.edu
}

\maketitle

\begin{abstract}
Image segmentation is a popular area of research in computer vision that has many applications in automated image processing. A recent technique called \emph{piecewise flat embeddings} (PFE) has been proposed for use in image segmentation; PFE transforms image pixel data into a lower dimensional representation where similar pixels are pulled close together and dissimilar pixels are pushed apart. This technique has shown promising results, but its original formulation is not computationally feasible for large images. We propose two improvements to the algorithm for computing PFE: first, we reformulate portions of the algorithm to enable various linear algebra operations to be performed in parallel; second, we propose utilizing an iterative linear solver (preconditioned conjugate gradient) to quickly solve a linear least-squares problem that occurs in the inner loop of a nested iteration. With these two computational improvements, we show on a publicly available image database that PFE can be sped up by an order of magnitude without sacrificing segmentation performance. Our results make this technique more practical for use on large data sets, not only for image segmentation, but for general data clustering problems.
\end{abstract}


\IEEEpeerreviewmaketitle

\section{Introduction}

Image segmentation is a popular area of research in computer vision and machine learning, as it is necessary for many applications such as automated visual recognition of objects in videos and photos and semantic image understanding. A recent method called \emph{piecewise flat embeddings} (PFE) \cite{yu2015pfe} performs image segmentation by creating a new pixel data representation (the \emph{embedding}) in which pixels that contain similar characteristics are close together, while dissimilar pixels are far apart. The main attribute that sets PFE apart from other embedding methods is the piecewise constant or "flat"' nature of the embeddings that makes clustering the pixels much easier than with other embeddings.

The PFE method relies on representing the image as a graph, with each pixel representing a vertex, and with similarities between pixels modeled by weighted edges between the vertices. As opposed to the well-known Laplacian Eigenmaps (LE) algorithm \cite{belkin2003led} that computes embeddings by minimizing a weighted $\ell_{2}$--norm of the differences between points in the new embedding, PFE minimizes a weighted $\ell_{1}$--norm subject to orthogonality constraints. This makes the embedding more difficult to compute than in LE, which reduces to a simple generalized eigenvalue problem. Yu et al. \cite{yu2015pfe} propose a numerical procedure to approximate the solution to PFE, involving a nested looping procedure that requires Bregman iterations \cite{goldstein2009sbm}. However, as described in \cite{yu2015pfe}, there are many computational limitations when solving the problem in this manner.

In this paper, we propose two improvements to the numerical procedure for approximating the PFE: reformulating various linear algebra operations defined in \cite{yu2015pfe} to use more compact matrices that enable much lower memory requirements, and utilizing an iterative linear solver (preconditioned conjugate gradient) as opposed to a direct solver to enable PFE to be computed on much larger graphs.  

\section{Piecewise Flat Embeddings - Background}

Suppose $\mathcal{X} = \left\{\mathbf{x}_{1},\ldots,\mathbf{x}_{n}\right\}$ is a set of data points in $\mathbb{R}^{m}$. Dimensionality reduction algorithms like PFE or LE aim to generate a new set of corresponding points $\mathcal{Y} = \left\{\mathbf{y}_{1},\ldots,\mathbf{y}_{n}\right\}$ in $\mathbb{R}^{d}$, where $d<<m$, so that inter-point distance relationships are preserved. To describe the relationships between the points in $\mathcal{X}$, we define a graph $\mathcal{G} = \left\langle \mathcal{V}, \mathcal{E} \right\rangle$ with vertex set $\mathcal{V} = \left\{v_{1},\ldots,v_{n}\right\}$ and edge set $\mathcal{E} \subseteq \mathcal{V}\times\mathcal{V}$, where each vertex $v_{i}$ corresponds to the point $\mathbf{x}_{i}$, and where the edge between vertices $v_{i}$ and $v_{j}$ is assigned a weight $w_{i,j}$. One common way to assign weights is according to the \emph{heat kernel}; i.e., $w_{i,j} = \exp\!\left(-\left\|\mathbf{x}_{i}-\mathbf{x}_{j}\right\|^{2}\!/2\sigma^{2}\right)$, where $\sigma$ is a parameter that can be selected by the user. Points in $\mathcal{X}$ that are nearby will correspond to edges in the graph with weights that approach one, whereas points in $\mathcal{X}$ that are far apart will correspond to edges with weights that approach zero.

Computing the PFE can be done by solving the constrained minimization problem:
\begin{align} \label{eq:pfe:orig}
	\min_{\mathbf{Y}}{\sum_{i,j}{w_{i,j}\!\left\|\mathbf{y}_{i}-\mathbf{y}_{j}\right\|_{1}}} 
	\quad \textrm{s.t.} \, \mathbf{Y^{T}DY} = \mathbf{I} \enspace ,
\end{align}
where $\mathbf{Y} = \left[\mathbf{y}_{1},\cdots,\mathbf{y}_{n}\right]^{\mathbf{T}}$ is the $n\times d$ matrix containing the new set of points in $\mathbb{R}^{d}$, $\mathbf{W}$ is the $n\times n$ \emph{weighted adjacency matrix} that contains the weights $w_{i,j}$ for each edge in the graph, and $\mathbf{D}$ is the \emph{degree matrix} of the graph (i.e., the diagonal matrix $\mathbf{D} = \textrm{diag}\!\left(\mathbf{d}\right)$, with $d_{i} = \sum_{j}{w_{i,j}}$).

The orthogonality constraint $\mathbf{Y^{T}DY} = \mathbf{I}$ is required so that we can avoid the trivial solution $\mathbf{Y} = \mathbf{0}$; however, it makes (\ref{eq:pfe:orig}) impossible to solve analytically. A numerical procedure for approximating its solution is presented in \cite{yu2015pfe}. This procedure relies on the Splitting Orthogonality Constraint (SOC) algorithm \cite{lai2014smo}. To carry out this procedure, Yu et al. \cite{yu2015pfe} define $\mathbf{P} = \mathbf{D}^{1/2}\mathbf{Y}$ and restate (\ref{eq:pfe:orig}) as:
\begin{align} \label{eq:pfe:orig:split}
	\min_{\mathbf{Y}}{\sum_{i,j}{w_{i,j}\!\left\|\mathbf{y}_{i}-\mathbf{y}_{j}\right\|_{1}}} 
	\quad \textrm{s.t.} \enspace \mathbf{D}^{1/2}\mathbf{Y} = \mathbf{P} \enspace , \enspace \mathbf{P^{T}P} = \mathbf{I} \enspace .
\end{align}
The SOC algorithm (Algorithm \ref{alg:SOC}) approximates the solution to (\ref{eq:pfe:orig:split}) by performing a Bregman iteration, as described in \cite{lai2014smo}.

\begin{algorithm}[t]
\caption{SOC Algorithm for Approximating (\ref{eq:pfe:orig:split})}\label{alg:SOC}
\begin{algorithmic}[1]
\Procedure{SOC}{$\mathbf{W},\mathbf{Y}^{\left(0\right)}$}
\State $\mathbf{D}\gets \textrm{diag}\!\left(\mathbf{W1}\right)$
\State $k = 0$, $\mathbf{P}^{\left(k\right)}\gets \mathbf{D}^{1/2}\mathbf{Y}^{\left(k\right)}$, $\mathbf{B}^{\left(k\right)}\gets \mathbf{0}_{n\times d}$
\Repeat
\State $\displaystyle \mathbf{Y}^{\left(k+1\right)} = \arg\min_{\mathbf{Y}}{\left(\sum_{ij}{w_{i,j}\!\left\|Y_{i}-Y_{j}\right\|_{1}} + \right.}$
\Statex \hfill $\displaystyle \tfrac{r}{2}\!\left. \left\|\mathbf{D}^{1/2}\mathbf{Y}-\mathbf{P}^{\left(k\right)}+\mathbf{P}^{\left(k\right)}\right\|^{2}_{2}\right)$
\State $\displaystyle \mathbf{P}^{\left(k+1\right)} = \arg\min_{\mathbf{P}}{\left\|\mathbf{P}\!-\!\left(\mathbf{D}^{1/2}\mathbf{Y}^{\left(k+1\right)}+\mathbf{B}^{\left(k\right)}\right)\right\|^{2}_{2}}$
\Statex \hfill s.t. $\displaystyle \mathbf{P^{T}P} = \mathbf{I} \quad\quad\quad\quad\quad\quad\quad\quad\quad\quad$
\State $\displaystyle \mathbf{B}^{\left(k+1\right)} = \mathbf{B}^{\left(k\right)} + \mathbf{D}^{1/2}\mathbf{Y}^{\left(k+1\right)} - \mathbf{P}^{\left(k+1\right)}$
\State $k\gets k+1$
\Until{convergence}
\State \textbf{return} $\mathbf{Y}^{\left(k\right)}$
\EndProcedure
\end{algorithmic}
\end{algorithm}

The update $\mathbf{P}^{\left(k+1\right)}$ (line 6 of Algorithm \ref{alg:SOC}) has a closed-form solution described in \cite{yu2015pfe} that relies on matrices computed in the previous step. That previous step in line 5 that updates $\mathbf{Y}^{\left(k+1\right)}$ is an $\ell_{1}$--norm minimization problem that can be solved by the Split Bregman algorithm \cite{goldstein2009sbm}, which transforms $\ell_{1}$--norm minimization problems into series of differentiable convex optimization problems.

To write the update $\mathbf{Y}^{\left(k+1\right)}$ in a manner that can be solved via Split Bregman, Yu et al. \cite{yu2015pfe} concatenate the columns of the matrices $\mathbf{Y}^{\left(k\right)}$, $\mathbf{P}^{\left(k\right)}$, and $\mathbf{B}^{\left(k\right)}$ into the vectors $\mathbf{Y}_{v}^{\left(k\right)}$, $\mathbf{P}_{v}^{\left(k\right)}$, and $\mathbf{B}_{v}^{\left(k\right)}$ respectively. They then define an $\left(n\!\left(n-1\right)\!/2\right)\times n$ sparse matrix $\mathbf{M}$ that contains only two nonzero entries per row. If the graph edges are ordered according to $\mathcal{E} = \left\{\left(v_{i_{k}},v_{j_{k}}\right), k = 1, \ldots , n\!\left(n-1\right)\!/2 \right\}$, then the only nonzero entries in the $k^{\textrm{th}}$ row of $\mathbf{M}$ are defined to be $M_{ki} = w_{i_{k},j_{k}}$ and $M_{kj} = -w_{i_{k},j_{k}}$. Next, a $\left(dn\!\left(n-1\right)\!/2\right)\times\left(dn\right)$ matrix $\mathbf{L}$ and a $\left(dn\right)\times\left(dn\right)$ matrix $\mathbf{\tilde{D}}$ are defined as follows:
\begin{align}
	\mathbf{L} &= \mathbf{I}_{d\times d}\otimes\mathbf{M} \enspace , \label{eq:L} \\
	\mathbf{\tilde{D}} &= \mathbf{I}_{d\times d}\otimes\mathbf{D} \enspace , \label{eq:Dtilde}
\end{align}
where $\otimes$ indicates Kronecker product. Using all of this notation allows step 5 in the SOC algorithm to be rewritten as:
\begin{align} 
	\mathbf{Y}_{v}^{\left(k+1\right)} &= \arg\min_{\mathbf{Y}_{v}} \left\|\mathbf{LY}_{v}\right\|_{1} + \nonumber \\
	&\quad\quad\quad\quad \frac{r}{2}\left\|\mathbf{\tilde{D}}^{1/2}\mathbf{Y}_{v}-\mathbf{P}^{\left(k\right)}_{v}+\mathbf{B}^{\left(k\right)}_{v}\right\|_{2}^{2} \enspace , \label{eq:SOCRewrite}
\end{align}
which can then be solved via Split-Bregman \cite{goldstein2009sbm}, as described in Algorithm \ref{alg:splitBregman}. The update $\mathbf{Y}_{v}^{\left(k,\ell+1\right)}$ in step 5 requires solving a linear least-squares problem.

\begin{algorithm}[t]
\caption{Split Bregman Algorithm for Approximating (\ref{eq:SOCRewrite})}\label{alg:splitBregman}
\begin{algorithmic}[1]
\Procedure{SplitBregman}{$\mathbf{M},\mathbf{D},\mathbf{P}^{\left(k\right)},\mathbf{B}^{\left(k\right)}$}
\State Construct $\mathbf{L}$ and $\mathbf{\tilde{D}}$ from (\ref{eq:L})--(\ref{eq:Dtilde}).
\State $\ell = 0$, $\mathbf{b}^{\ell}\gets \mathbf{0}_{\left(dn\left(n-1\right)\!/2\right)\times 1}$, $\mathbf{d}^{\ell}\gets \mathbf{0}_{\left(dn\left(n-1\right)\!/2\right)\times 1}$
\Repeat
\State $\displaystyle \mathbf{Y}_{v}^{\left(k,\ell + 1\right)} = \arg\min_{\mathbf{Y}_{v}}{\left( \tfrac{\lambda}{2}\!\left\|\mathbf{LY}_{v}+\mathbf{b}^{\ell}-\mathbf{d}^{\ell}\right\|_{2}^{2} + \right.}$
\Statex \hfill $\displaystyle +\tfrac{r}{2}\!\left. \left\|\mathbf{\tilde{D}}^{1/2}\mathbf{Y}_{v}-\mathbf{P}^{\left(k\right)}_{v}+\mathbf{B}^{\left(k\right)}_{v}\right\|^{2}_{2}\right)$
\State $\displaystyle \mathbf{d}^{\ell + 1} = \textrm{Shrink}\!\left(\mathbf{LY}^{\left(k,\ell + 1\right)}_{v}+\mathbf{b}^{\ell},1/\lambda\right)$
\State $\displaystyle \mathbf{b}^{\ell + 1} = \mathbf{b}^{\ell} + \mathbf{LY}^{\left(k,\ell + 1\right)}_{v} - \mathbf{d}^{\ell + 1}$
\State $\ell\gets \ell+1$
\Until{convergence}
\State \textbf{return} $\mathbf{Y}^{\left(k,\ell\right)}$
\EndProcedure
\Procedure{Shrink}{$\mathbf{y}$,$\gamma$}
\State $z_{i} = \textrm{sign}\!\left(y_{i}\right)\!\cdot\!\max\!\left(\left|y_{i}\right| - \gamma, 0\right)$, $i = 1, \ldots, \textrm{numel}\!\left(\mathbf{z}\right)$
\State \textbf{return} $\mathbf{z}$
\EndProcedure
\end{algorithmic}
\end{algorithm}

\section{Efficiently Computing PFE}

Suppose we would like to apply the PFE method to segment a $128\times 128$ pixel image into $64$ segments, and we create a graph where each vertex corresponds to a single pixel, and the only edges having nonzero weights connect vertices representing pixels that are within each others' $4$-nearest neighbors. The sparse weighted adjacency matrix $\mathbf{W}$ would require approximately $0.5$ MB to store using double precision floating point values, and the matrix $\mathbf{Y}$ would require $8$ MB. For the Split-Bregman algorithm to be applied, the sparse matrices $\mathbf{M}$, $\mathbf{L}$, and $\mathbf{\tilde{D}}$ need to be computed; $\mathbf{M}$ will be $134,\!209,\!536\times 16,\!384$ with $131,\!072$ non-zero entries ($1$ MB), $\mathbf{L}$ will be $8,\!589,\!410,\!304\times 1,\!048,\!576$ with $8,\!388,\!608$ non-zero entries ($64$ MB), and $\mathbf{D}$ will be $1,\!048,\!576\times 1,\!048,\!576$ with non-zero entries on the main diagonal ($8$ MB). 

Even for this small image with modest neighborhood structure, the amount of memory needed to store these matrices is large. The simple step of multiplying $\mathbf{L}$ by the vector $\mathbf{Y}_{v}$ in every Split-Bregman iteration will require significant computational effort. What becomes computationally prohibitive is step 5 of Algorithm \ref{alg:splitBregman}. The normal equations for this linear least squares problem are:
\begin{align} \label{eq:normal}
	\left[\tfrac{\lambda}{2}\mathbf{L^{T}L} + \tfrac{r}{2}\mathbf{\tilde{D}}\right]\!\mathbf{Y}_{v}^{\left(k,\ell+1\right)} &= \tfrac{\lambda}{2}\mathbf{L^{T}q}_{1} + \tfrac{r}{2}\mathbf{\tilde{D}}^{1/2}\mathbf{q}_{2} \enspace , 
\end{align}
where $\mathbf{q}_{1} = \mathbf{d}^{\ell}-\mathbf{b}^{\ell}$ and $\mathbf{q}_{2} = \mathbf{B}^{\left(k\right)}_{v}-\mathbf{P}^{\left(k\right)}_{v}$. It would be unwise to invert $\tfrac{\lambda}{2}\mathbf{L^{T}L} + \tfrac{r}{2}\mathbf{\tilde{D}}$ to solve (\ref{eq:normal}); for our example, the inverse would be a dense $1,\!048,\!576\times 1,\!048,\!576$ matrix, requiring $8$ TB of storage. Even attempting to use a direct solver for (\ref{eq:normal}) would require too much memory to be feasible, because although $\tfrac{\lambda}{2}\mathbf{L^{T}L} + \tfrac{r}{2}\mathbf{\tilde{D}}$ is sparse and banded, its bands are far away from the main diagonal.

In order to combat these problems in computing PFE, we first note that if we define the function $\textrm{vec}:\mathbb{R}^{s\times t}\rightarrow\mathbb{R}^{st}$ that "unwraps" a matrix $\mathbf{Z} = \left[\mathbf{z}_{1},\ldots,\mathbf{z}_{t}\right]$ into the vector $\textrm{vec}\!\left(\mathbf{Z}\right) = \left[\mathbf{z}_{1}^{\mathbf{T}},\ldots,\mathbf{z}_{t}^{\mathbf{T}}\right]^{\mathbf{T}}$, then we can write $\mathbf{Y}_{v}^{\left(k\right)} = \textrm{vec}\!\left(\mathbf{Y}^{\left(k\right)}\right)$, $\mathbf{P}_{v}^{\left(k\right)} = \textrm{vec}\!\left(\mathbf{P}^{\left(k\right)}\right)$, and $\mathbf{B}_{v}^{\left(k\right)} = \textrm{vec}\!\left(\mathbf{B}^{\left(k\right)}\right)$. Using this notation, we can write:
\begin{align}
	\mathbf{LY}_{v} &= \textrm{vec}\!\left(\mathbf{MY}\right) \enspace , \label{eq:LYv} \\
	\mathbf{\tilde{D}}^{1/2}\mathbf{Y}_{v} &= \textrm{vec}\!\left(\mathbf{D}^{1/2}\mathbf{Y}\right) \enspace . \label{eq:DhalfYv}
\end{align}
Eq. (\ref{eq:LYv}) allows us to rewrite steps 6--7 of the Split-Bregman algorithm as:
\begin{align}
	\mathbf{d}^{\ell + 1} &= \textrm{Shrink}\!\left(\textrm{vec}\!\left(\mathbf{MY}^{\left(k,\ell + 1\right)}\right)+\mathbf{b}^{\ell},1/\lambda\right) \enspace , \label{eq:newdell} \\
	\mathbf{b}^{\ell + 1} &= \mathbf{b}^{\ell} + \textrm{vec}\!\left(\mathbf{MY}^{\left(k,\ell + 1\right)}\right) - \mathbf{d}^{\ell + 1} \enspace , \label{eq:newbell}
\end{align}
which reduces the computation required for these steps by a factor of $d$. Furthermore, we can write:
\begin{align}
	\mathbf{L^{T}LY}_{v} &= \textrm{vec}\!\left(\mathbf{M^{T}MY}\right) \enspace , \label{eq:LTLYv} \\
	\mathbf{\tilde{D}Y}_{v} &= \textrm{vec}\!\left(\mathbf{DY}\right) \enspace , \label{eq:DYv}
\end{align}
and therefore, (\ref{eq:normal}) can be expressed alternatively as:
\begin{align} \label{eq:normalEfficient}
	\left[\tfrac{\lambda}{2}\mathbf{M^{T}M} + \tfrac{r}{2}\mathbf{D}\right]\!\mathbf{Y}^{\left(k,\ell+1\right)} &= \tfrac{\lambda}{2}\mathbf{M^{T}Q}_{1} + \tfrac{r}{2}\mathbf{D}^{1/2}\mathbf{Q}_{2} \enspace , 
\end{align}
where $\mathbf{d}^{\ell}-\mathbf{b}^{\ell} = \textrm{vec}\!\left(\mathbf{Q}_{1}\right)$ and $\mathbf{Q}_{2} = \mathbf{B}^{\left(k\right)}-\mathbf{P}^{\left(k\right)}$. This provides our first improvement in efficiency in computing PFE: instead of solving the single $\left(dn\right)\times\left(dn\right)$ system of equations (\ref{eq:normal}), we \emph{simultaneously} solve the $d$ different $n\times n$ systems of equations in (\ref{eq:normalEfficient}). In fact, if we replace steps 5, 6, and 7 in the Split-Bregman algorithm with (\ref{eq:normalEfficient}), (\ref{eq:newdell}), and (\ref{eq:newbell}) respectively, we see that the large matrices $\mathbf{L}$ and $\mathbf{\tilde{D}}$ never have to be explicitly formed. 

Secondly, we note that even though we have reduced the memory requirements of the system matrix in (\ref{eq:normalEfficient}) by a factor of $d$ over (\ref{eq:normal}), we found by experimentation that it is still infeasible to solve (\ref{eq:normalEfficient}) by explicitly inverting the system matrix or by invoking a direct solver. For this reason, we approximate the solution to (\ref{eq:normalEfficient}) via the preconditioned conjugate gradient (PCG) method \cite{barrett1994tsl} with an incomplete Cholesky preconditioner.

\section{Experiments}

In order to investigate how well our efficient PFE method performs on an image segmentation task, we replicate one of the experiments in \cite{yu2015pfe} that uses the $200$ training images from the BSDS500 dataset \cite{arbelaez2011cdh}, computes PFE, and performs segmentation by $k$--means clustering on the embeddings. We follow the two-stage approach of \cite{yu2015pfe} that first carries out the nested Bregman iteration (Stage I) and then relaxes the orthogonality constraint and only carries out the Split-Bregman algorithm (Stage II). In Stage I, a maximum of $10$ SOC iterations and $5$ Split-Bregman iterations are performed, and in Stage II, a maximum of $100$ Split-Bregman iterations are performed. The parameters $\lambda$ and $r$ are selected as in \cite{yu2015pfe}.

\begin{figure*}[t]
	\begin{center}
		\begin{tabular}{P{1in} P{1in} P{1in} P{1in} P{1in} P{1in}}
				\includegraphics[width=1.1in]{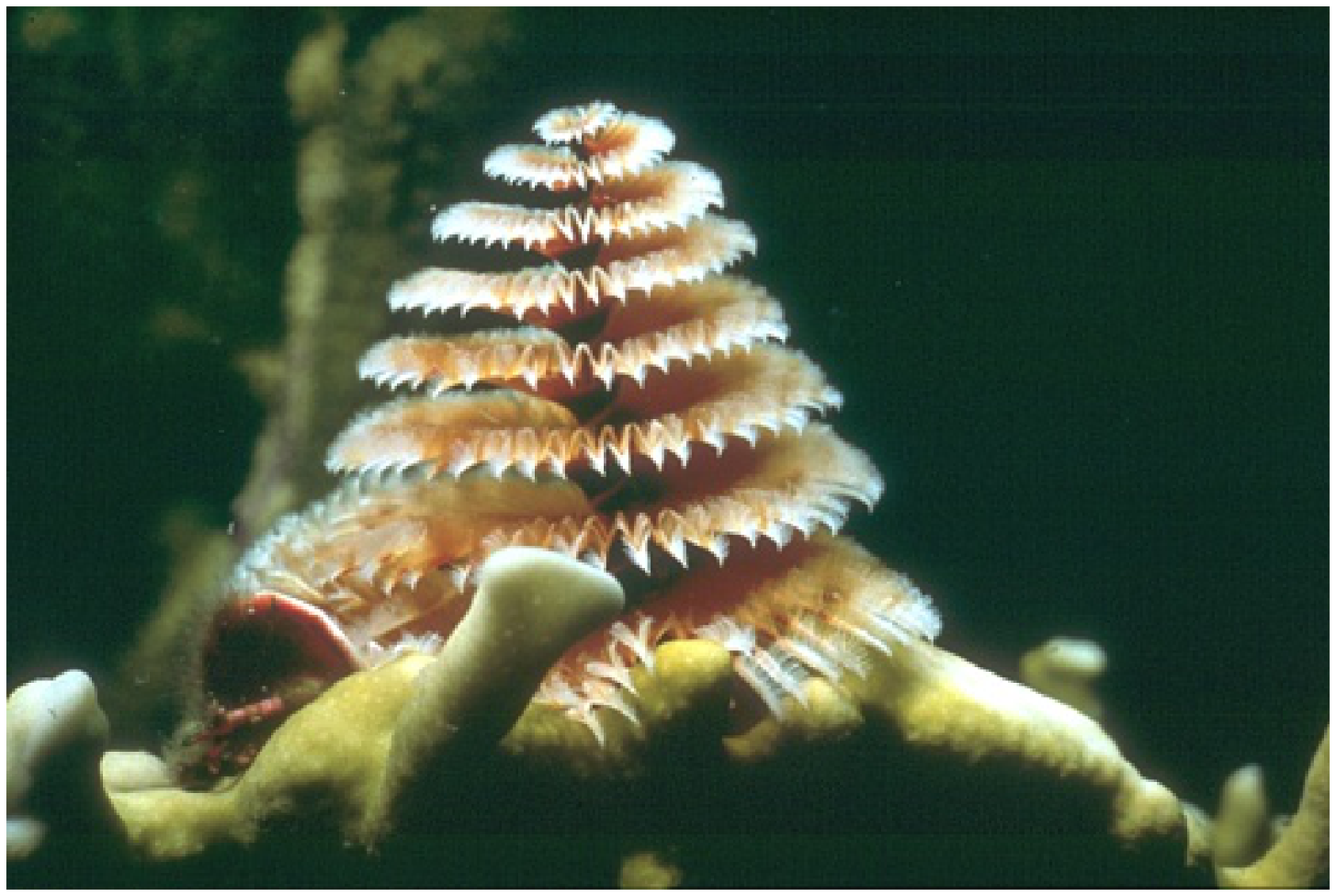} &
				\includegraphics[width=1.1in]{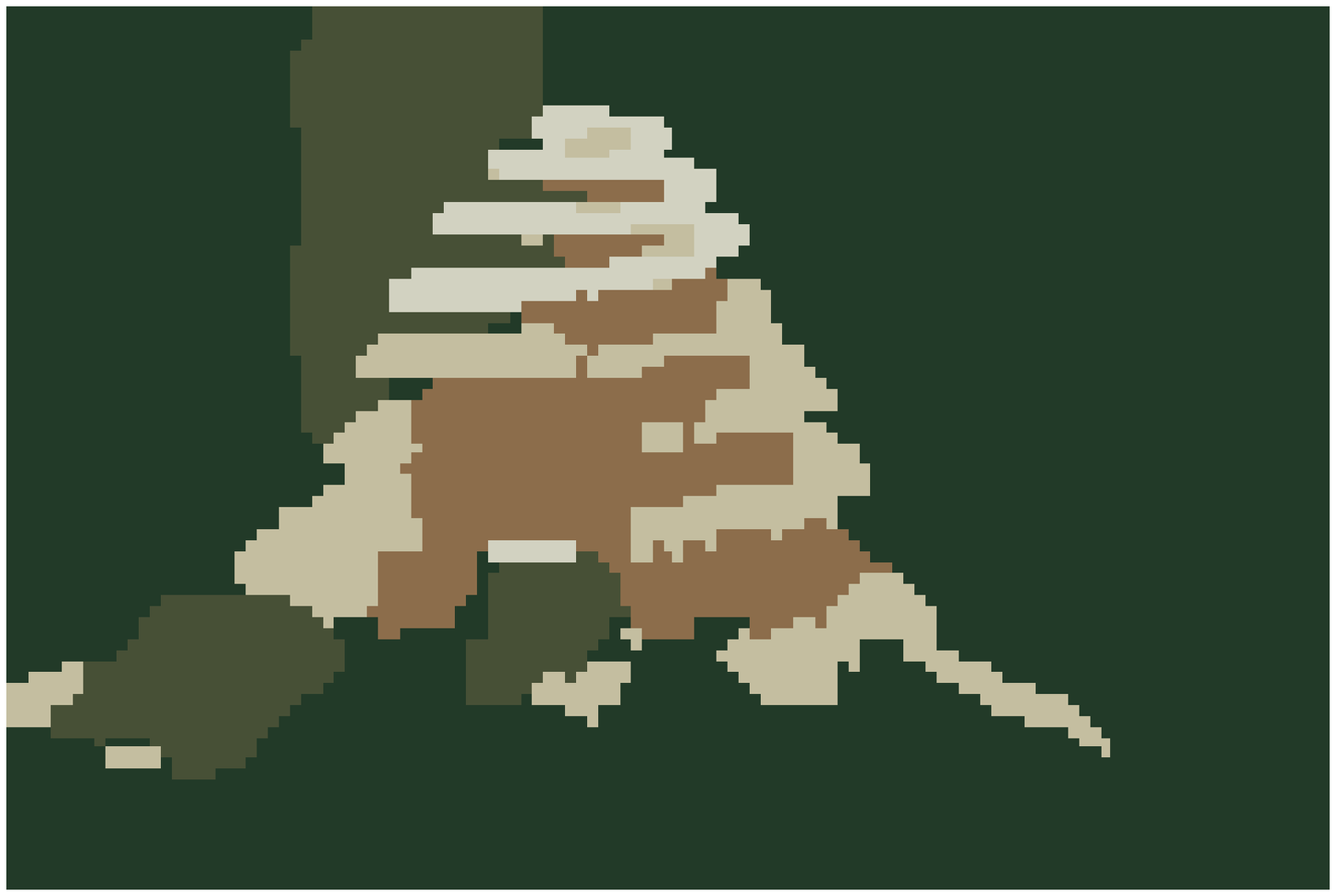} &
				\includegraphics[width=1.1in]{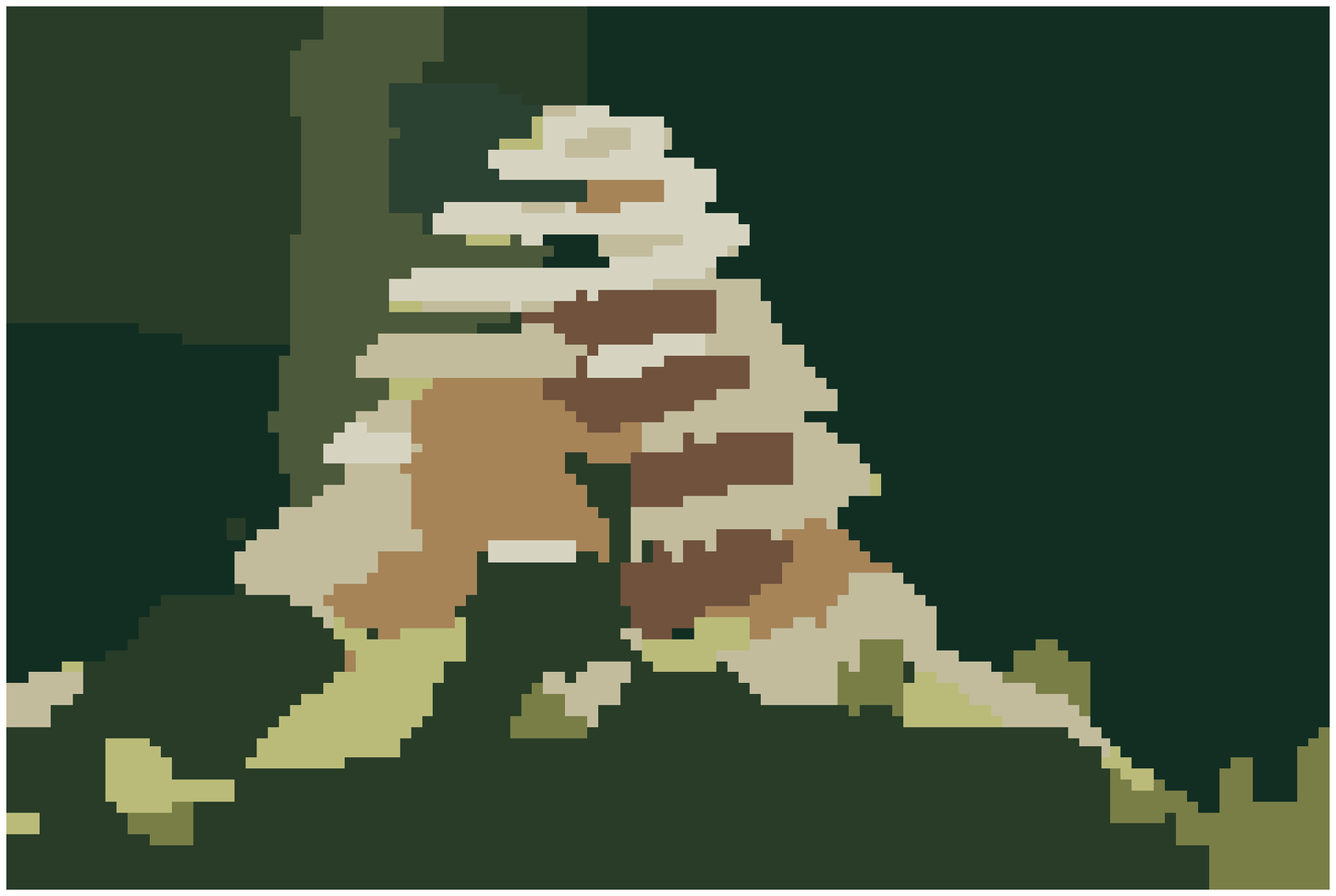} &
				\includegraphics[width=1.1in]{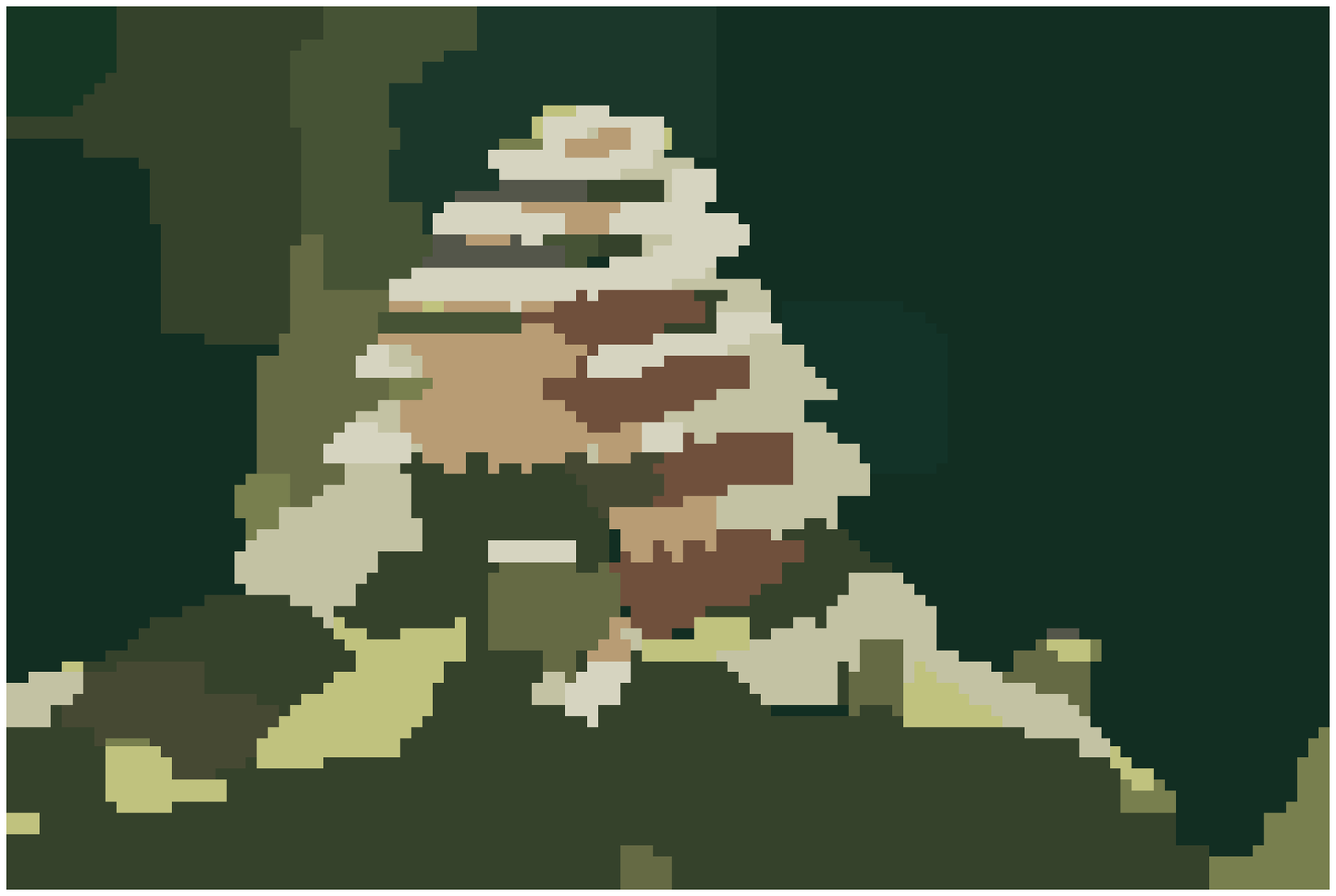} &
				\includegraphics[width=1.1in]{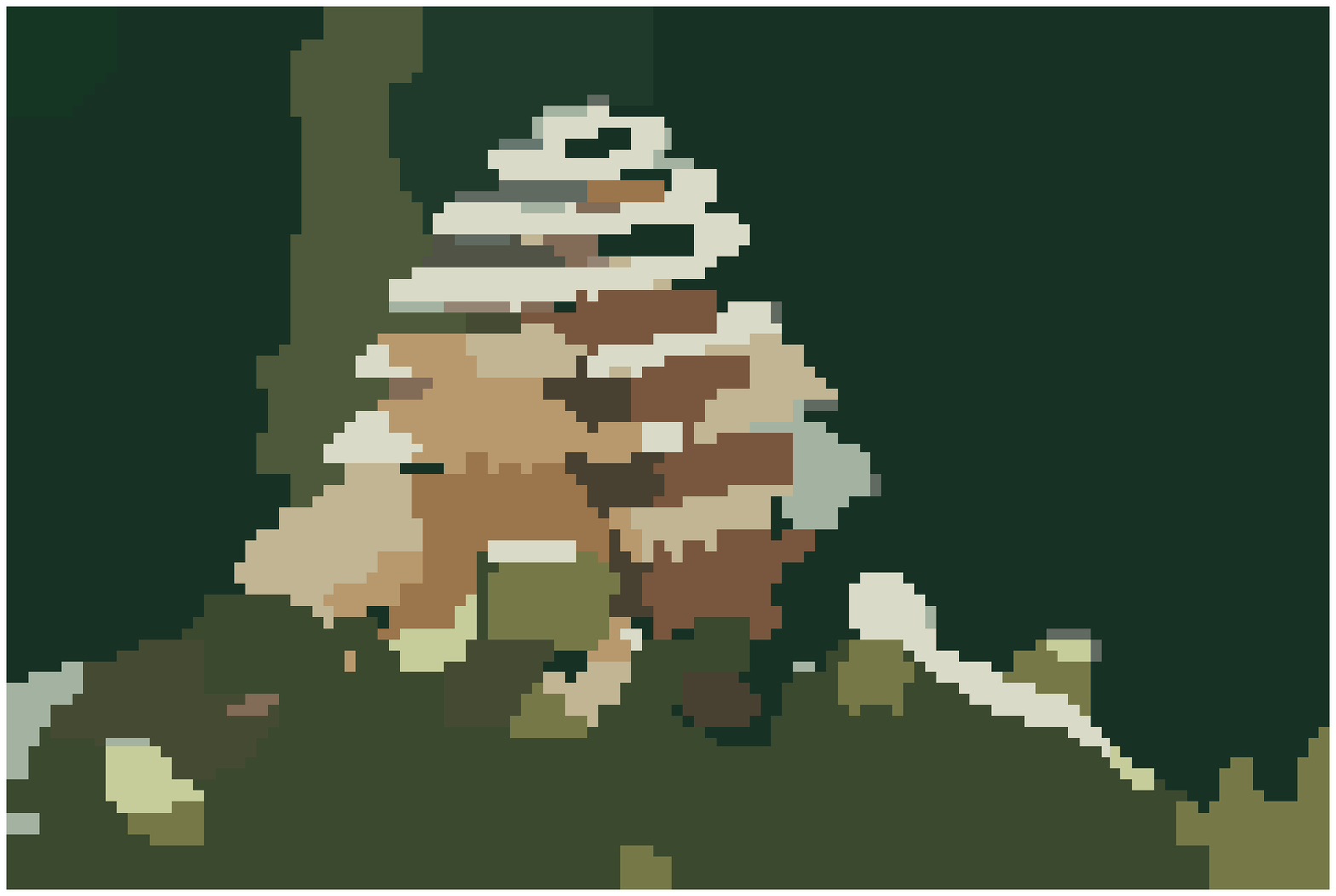} &
				\includegraphics[width=1.1in]{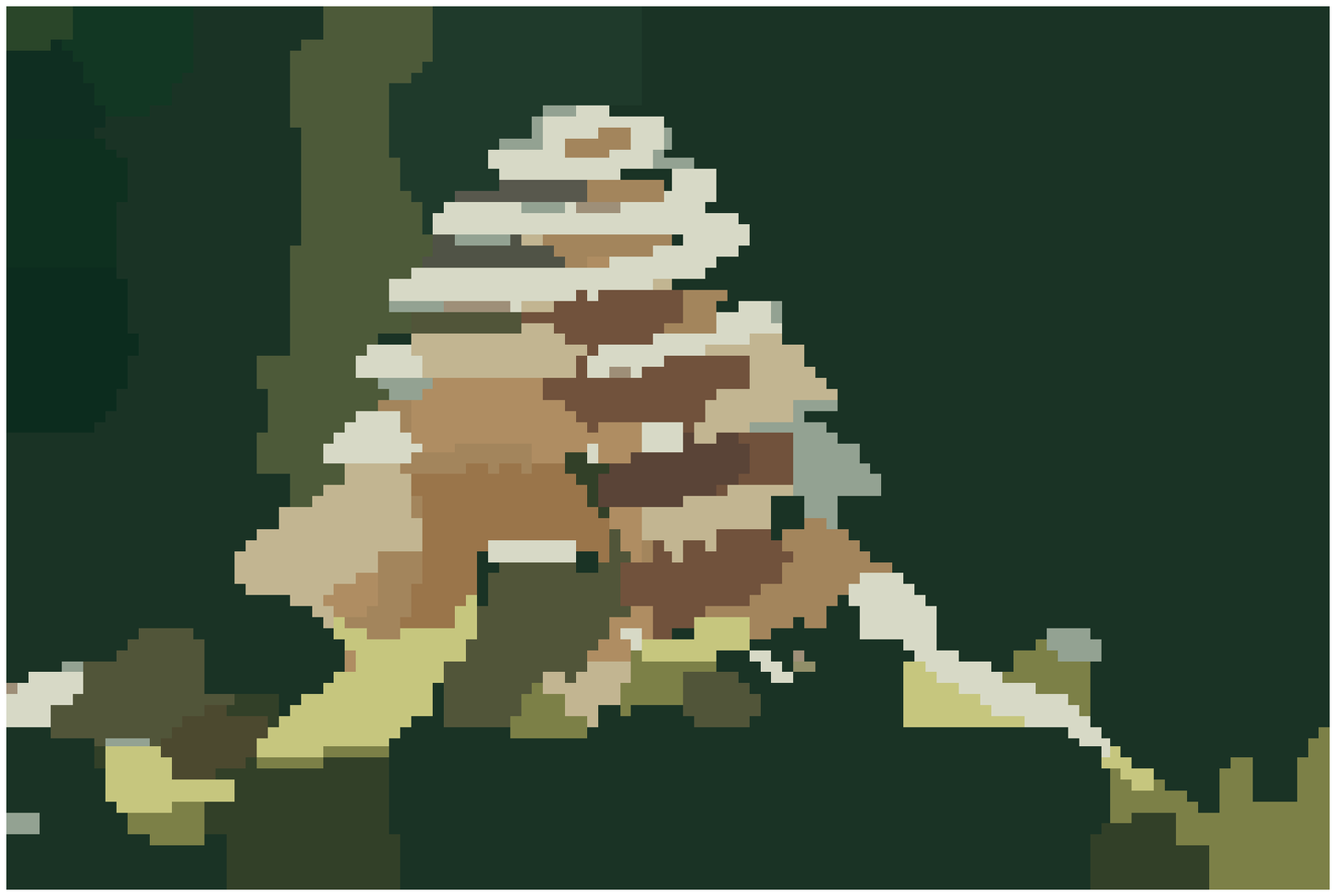} \\
				\includegraphics[width=1.1in]{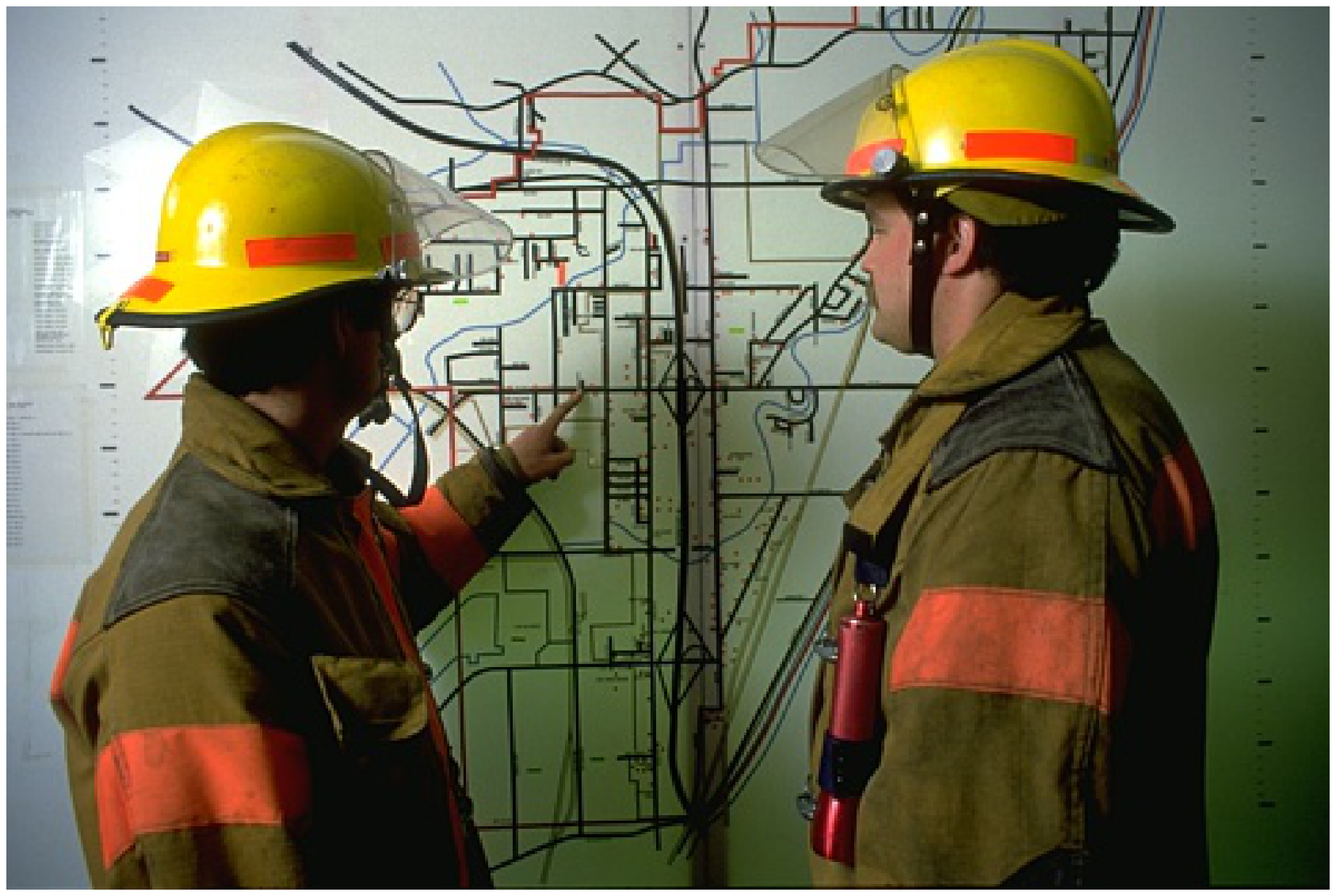} &
				\includegraphics[width=1.1in]{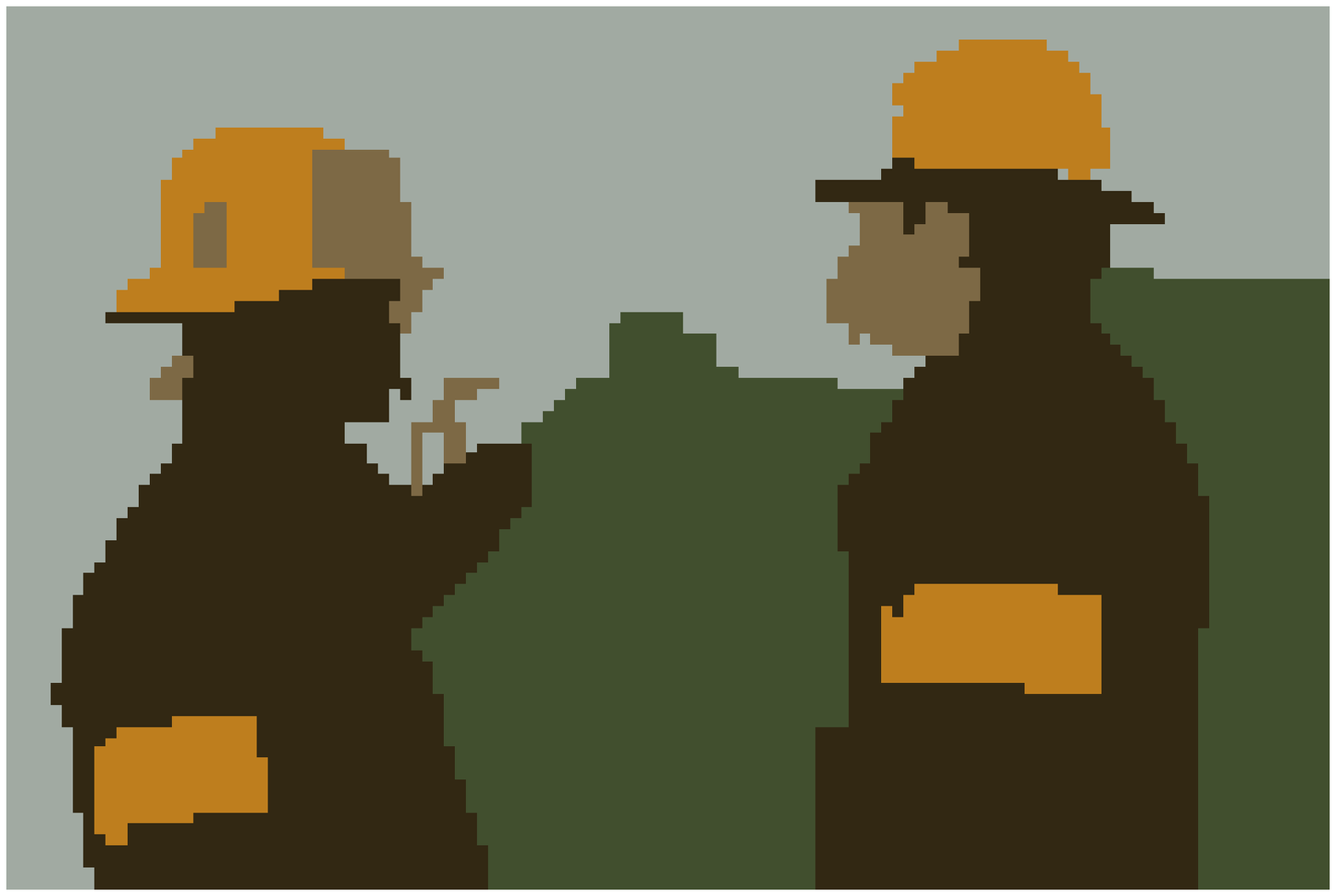} &
				\includegraphics[width=1.1in]{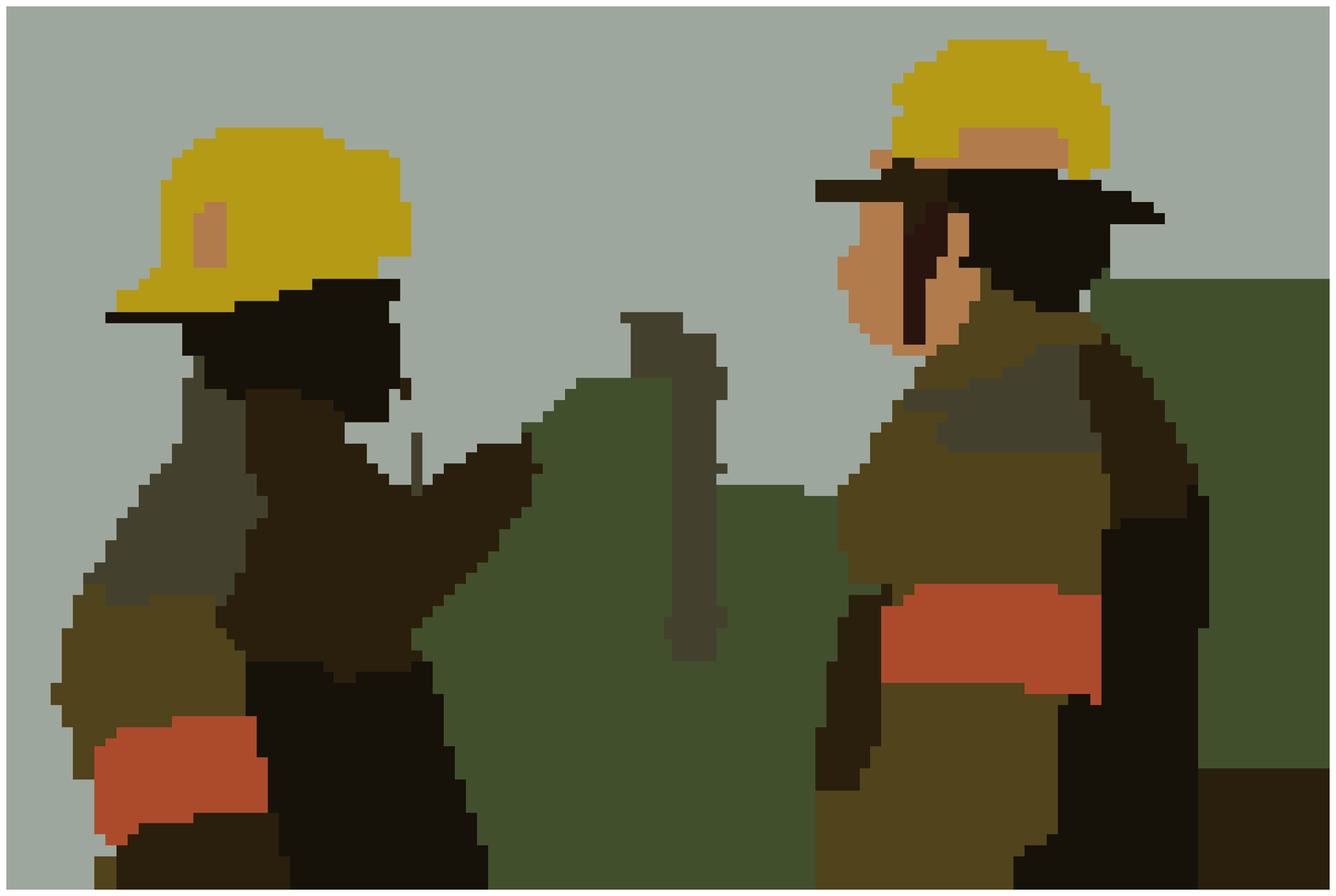} &
				\includegraphics[width=1.1in]{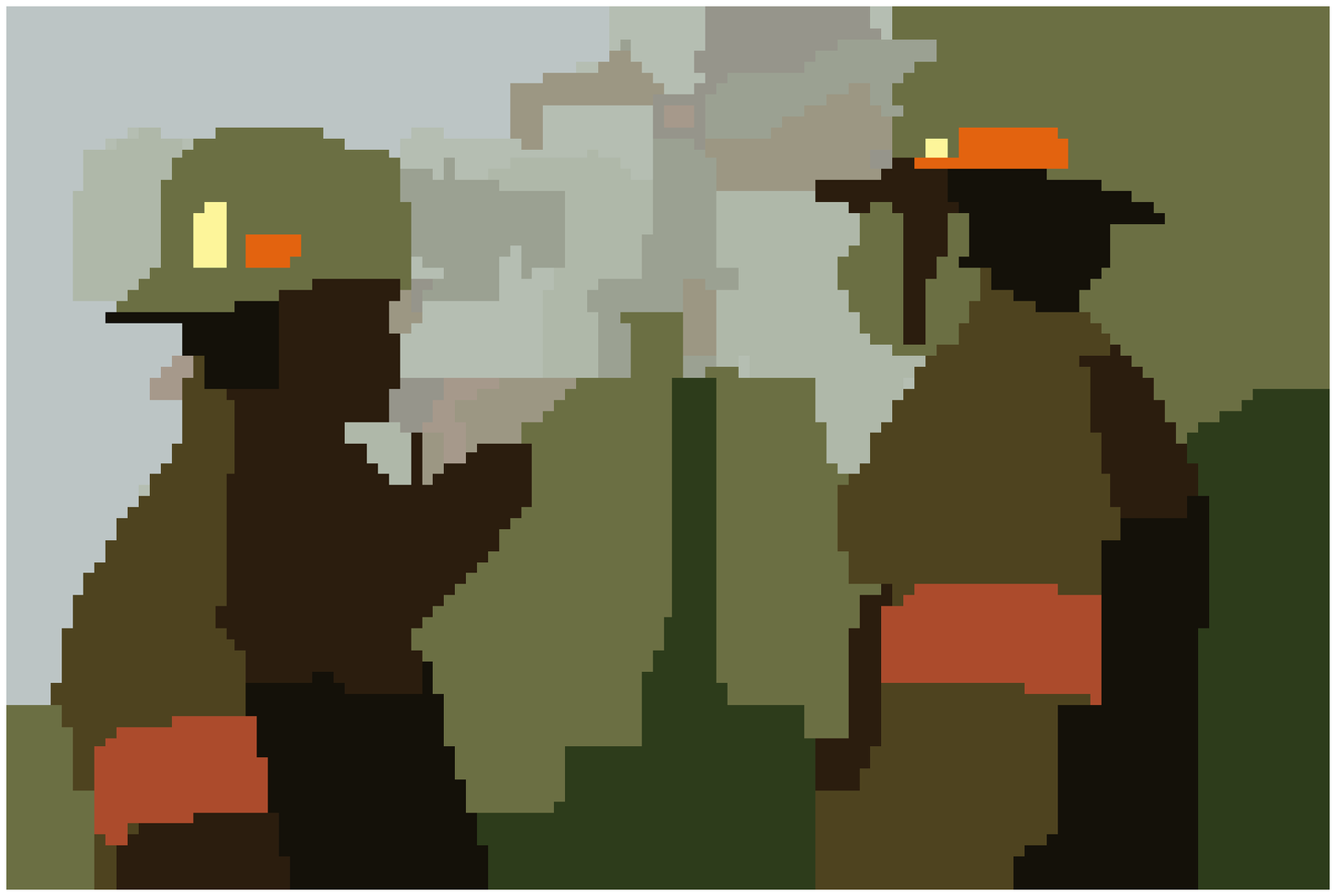} &
				\includegraphics[width=1.1in]{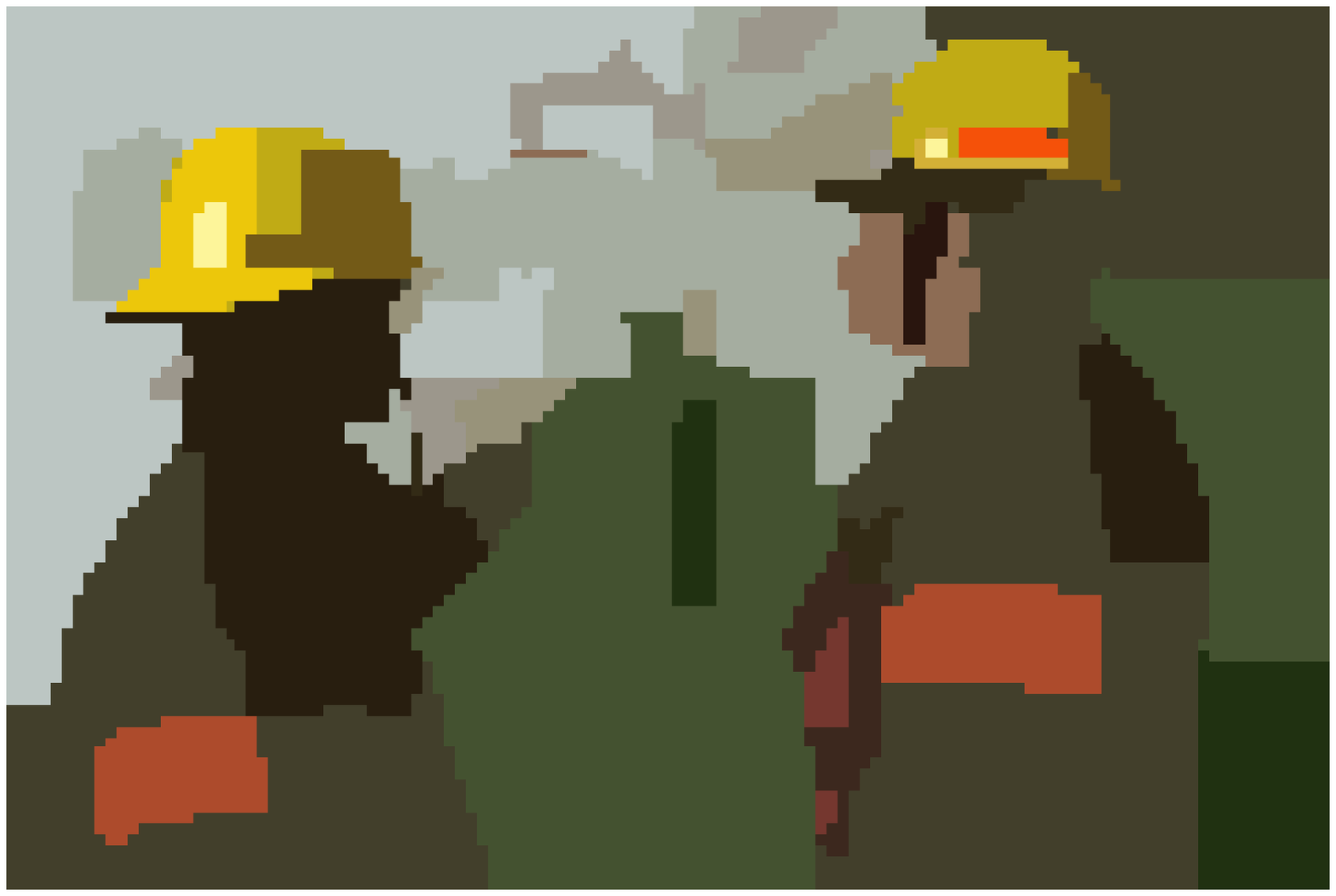} &
				\includegraphics[width=1.1in]{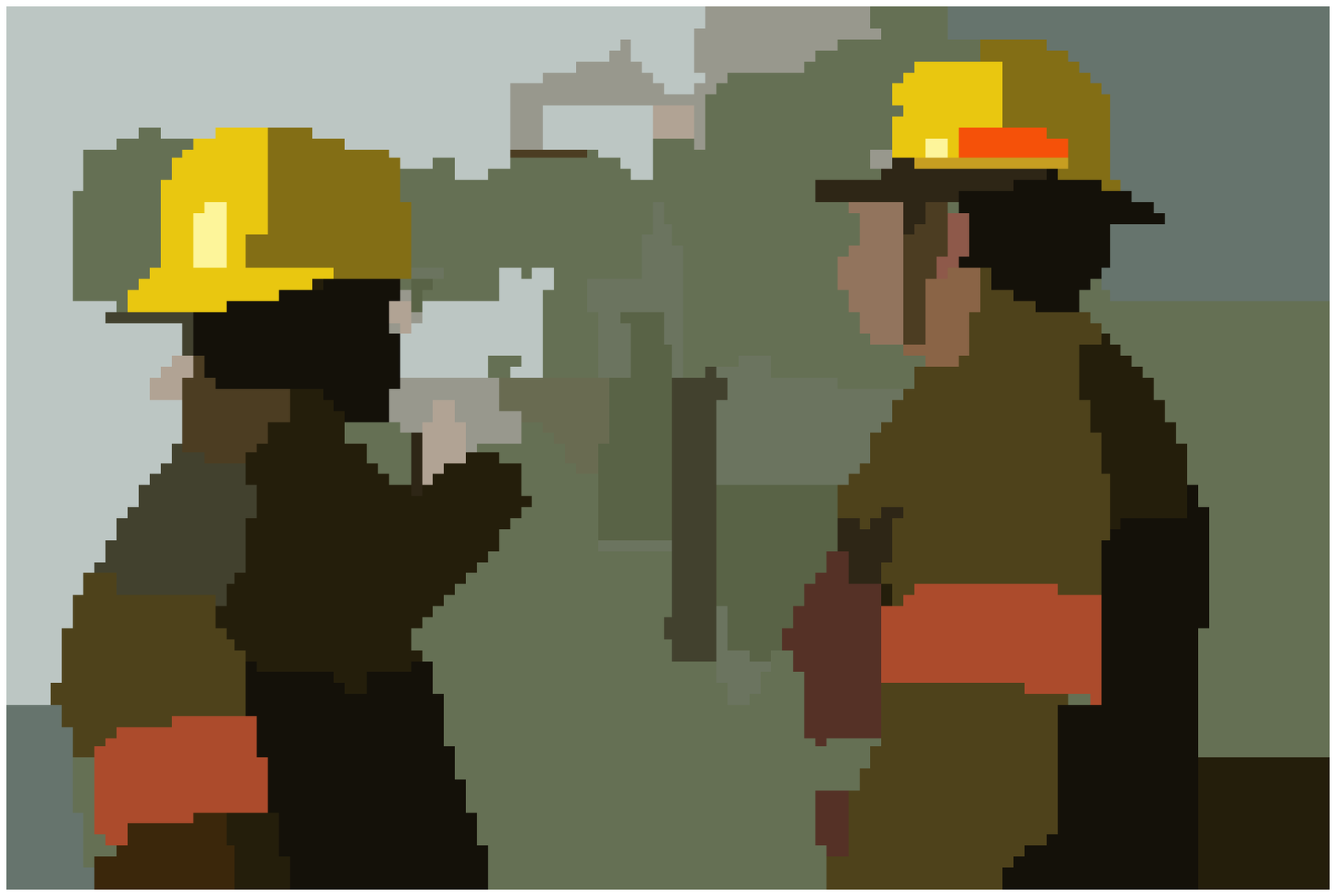} \\
				\includegraphics[width=1.1in]{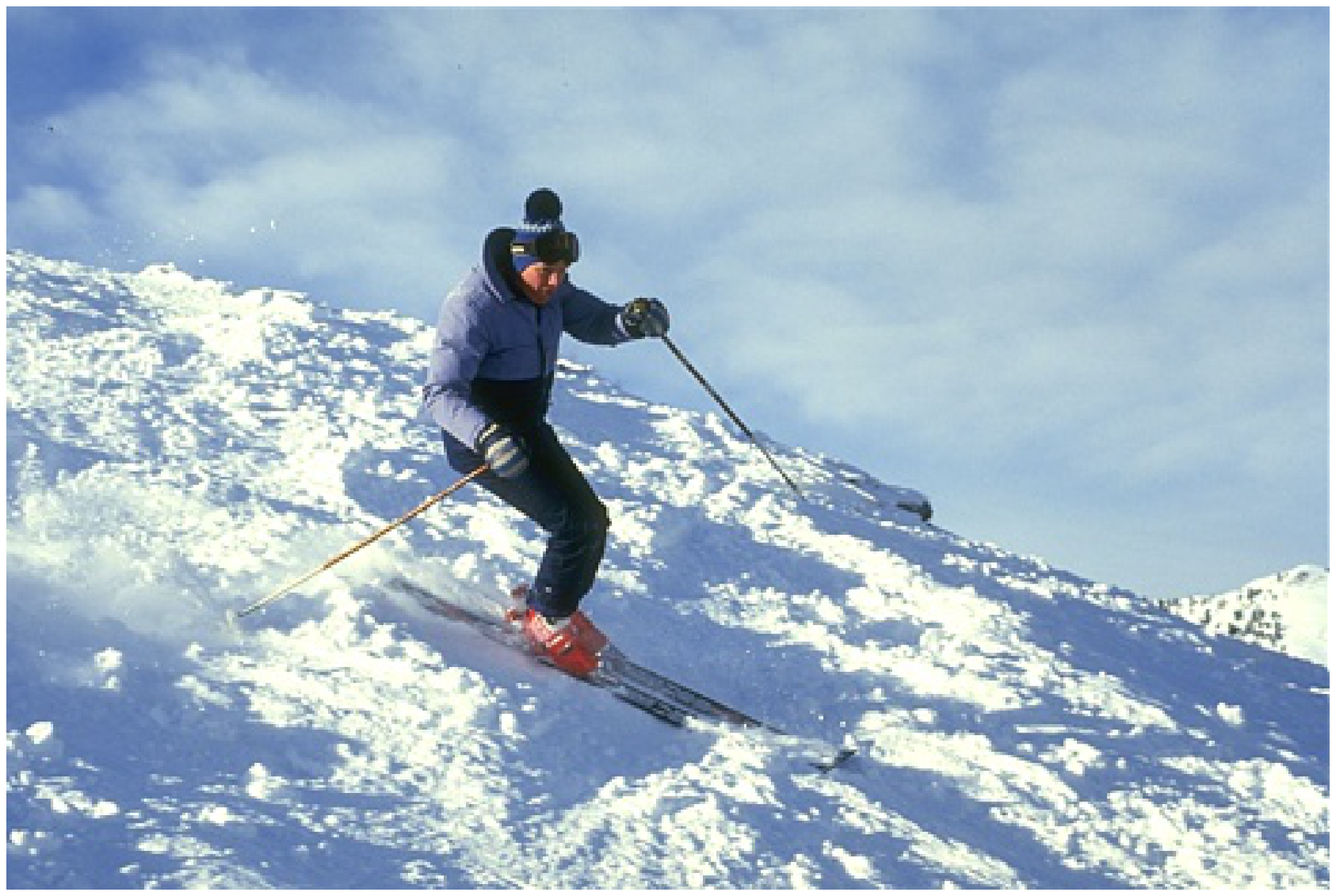} &
				\includegraphics[width=1.1in]{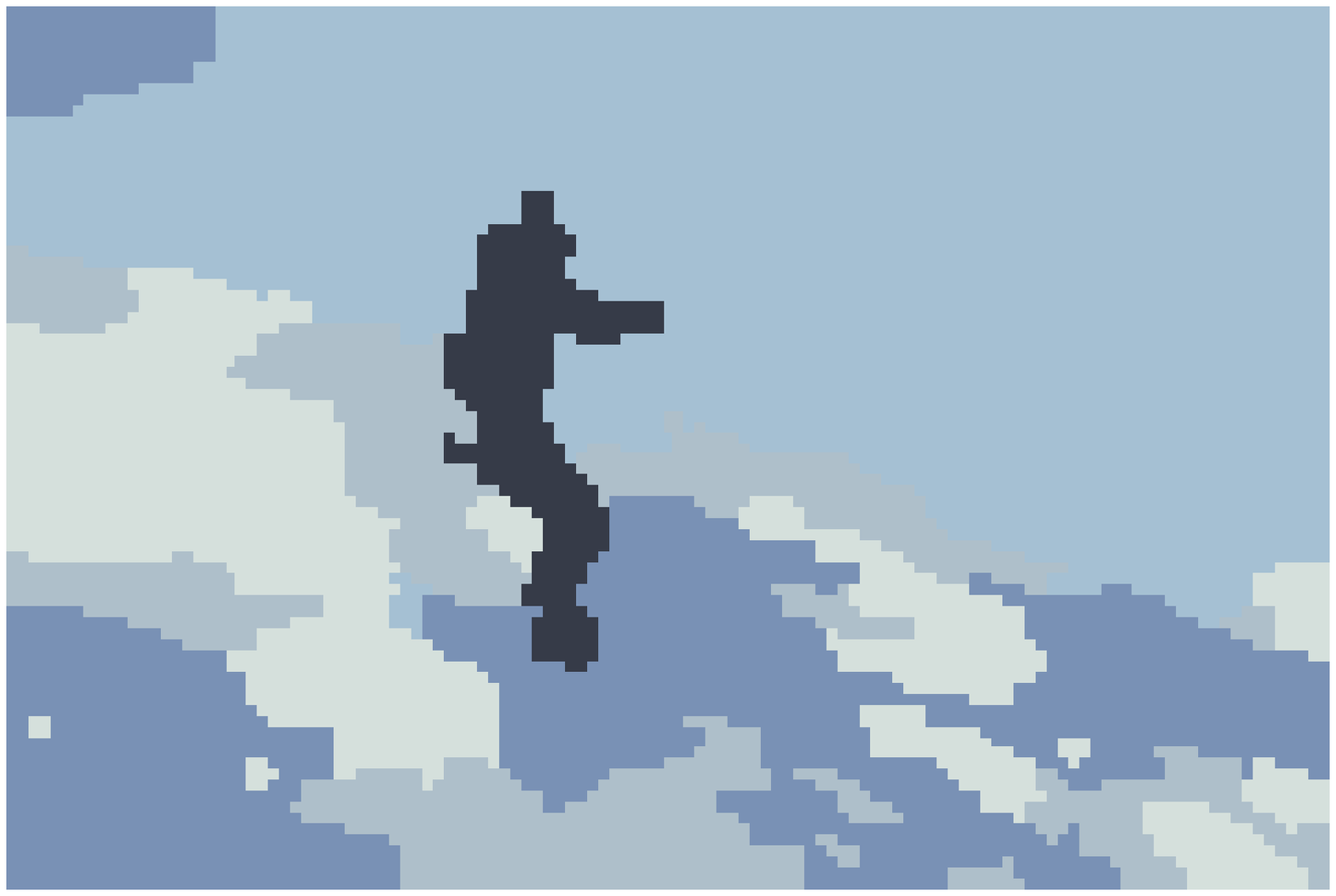} &
				\includegraphics[width=1.1in]{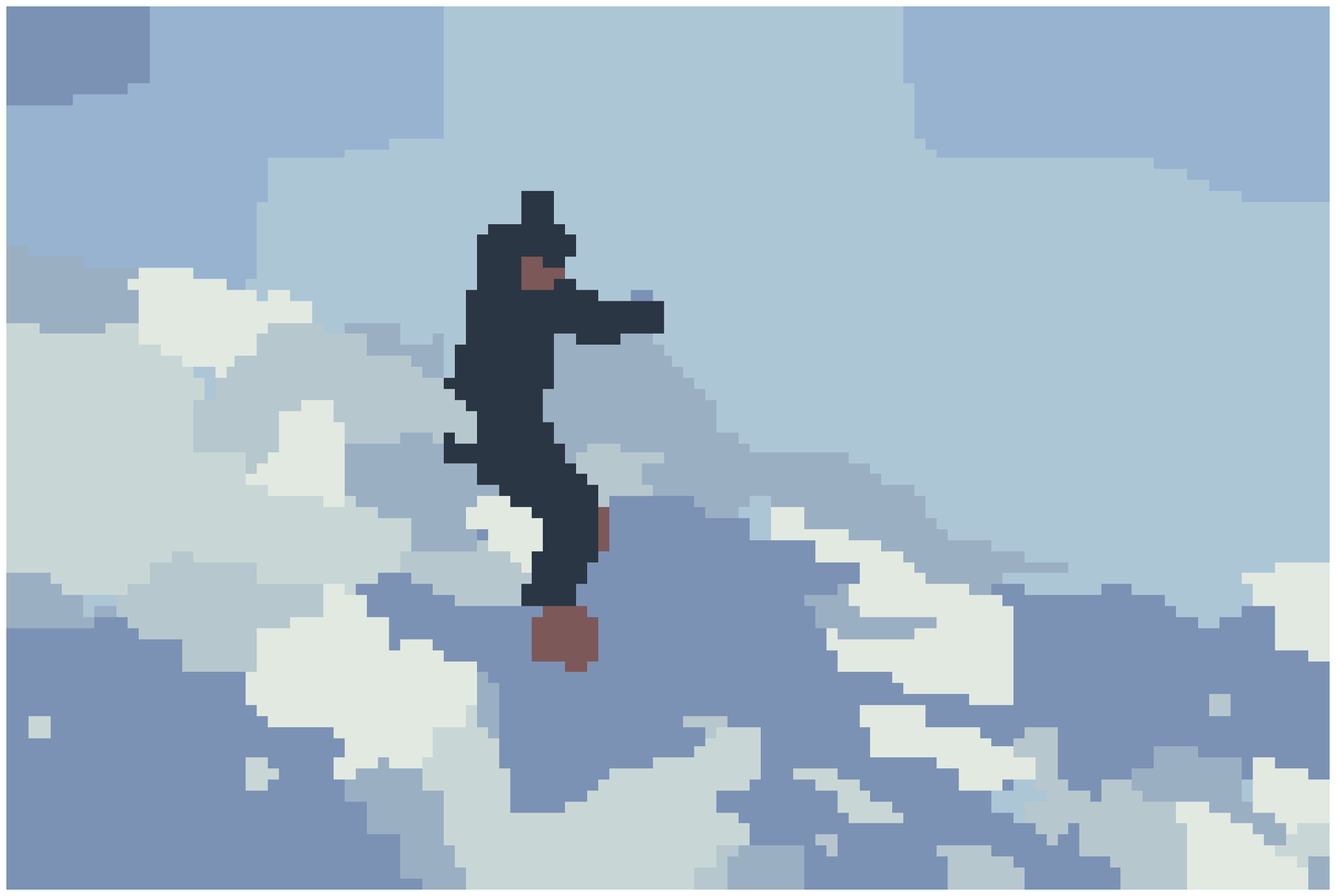} &
				\includegraphics[width=1.1in]{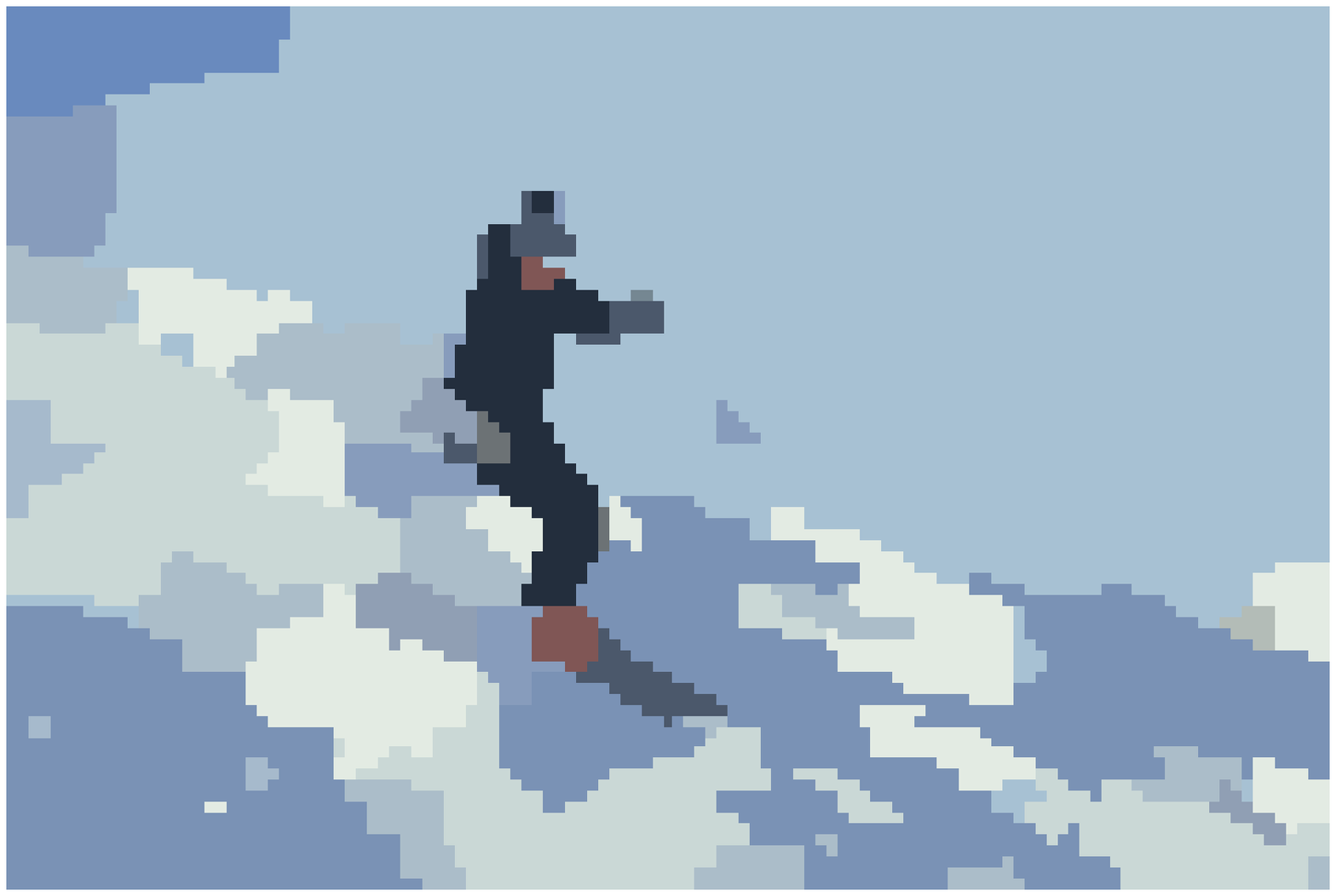} &
				\includegraphics[width=1.1in]{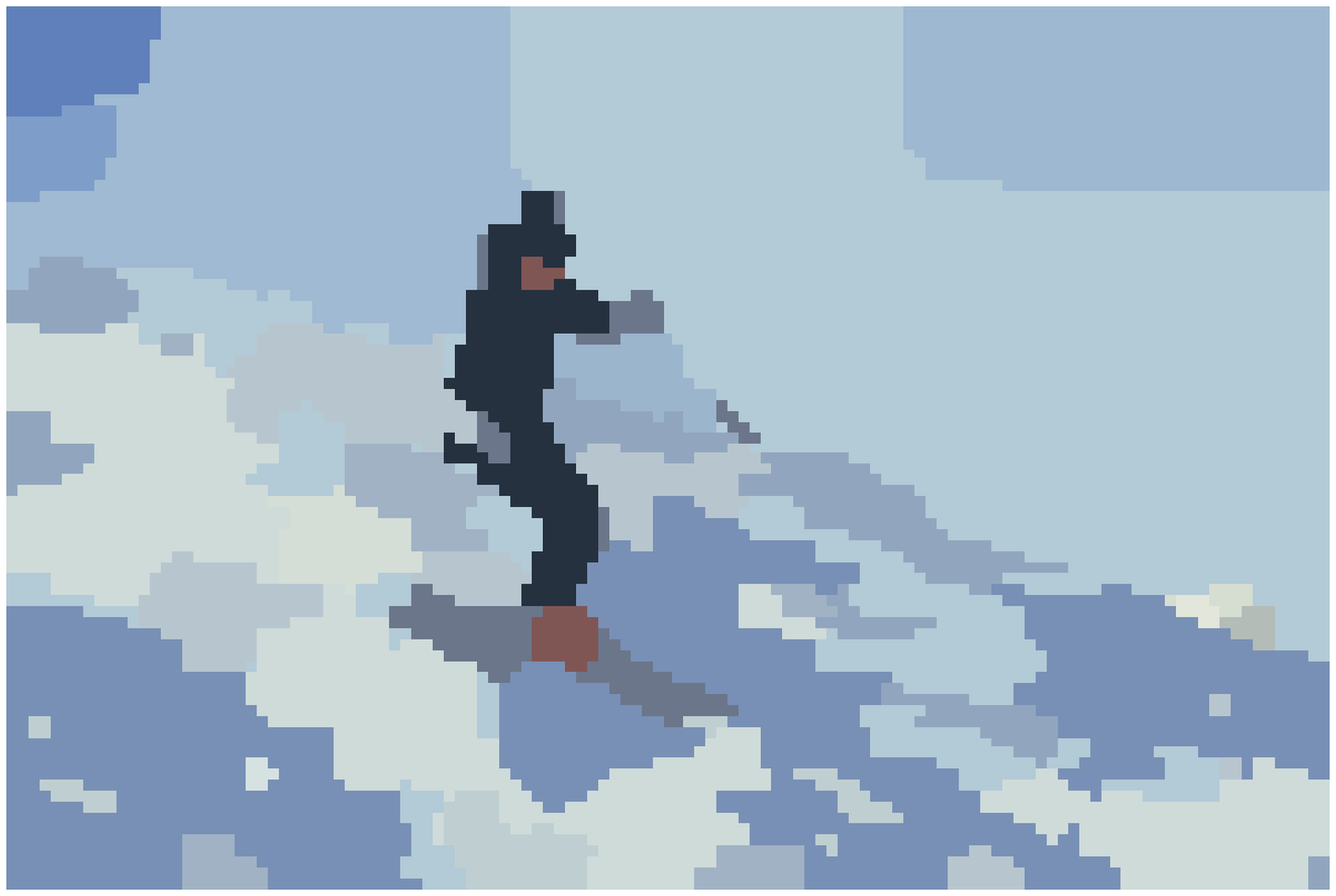} &
				\includegraphics[width=1.1in]{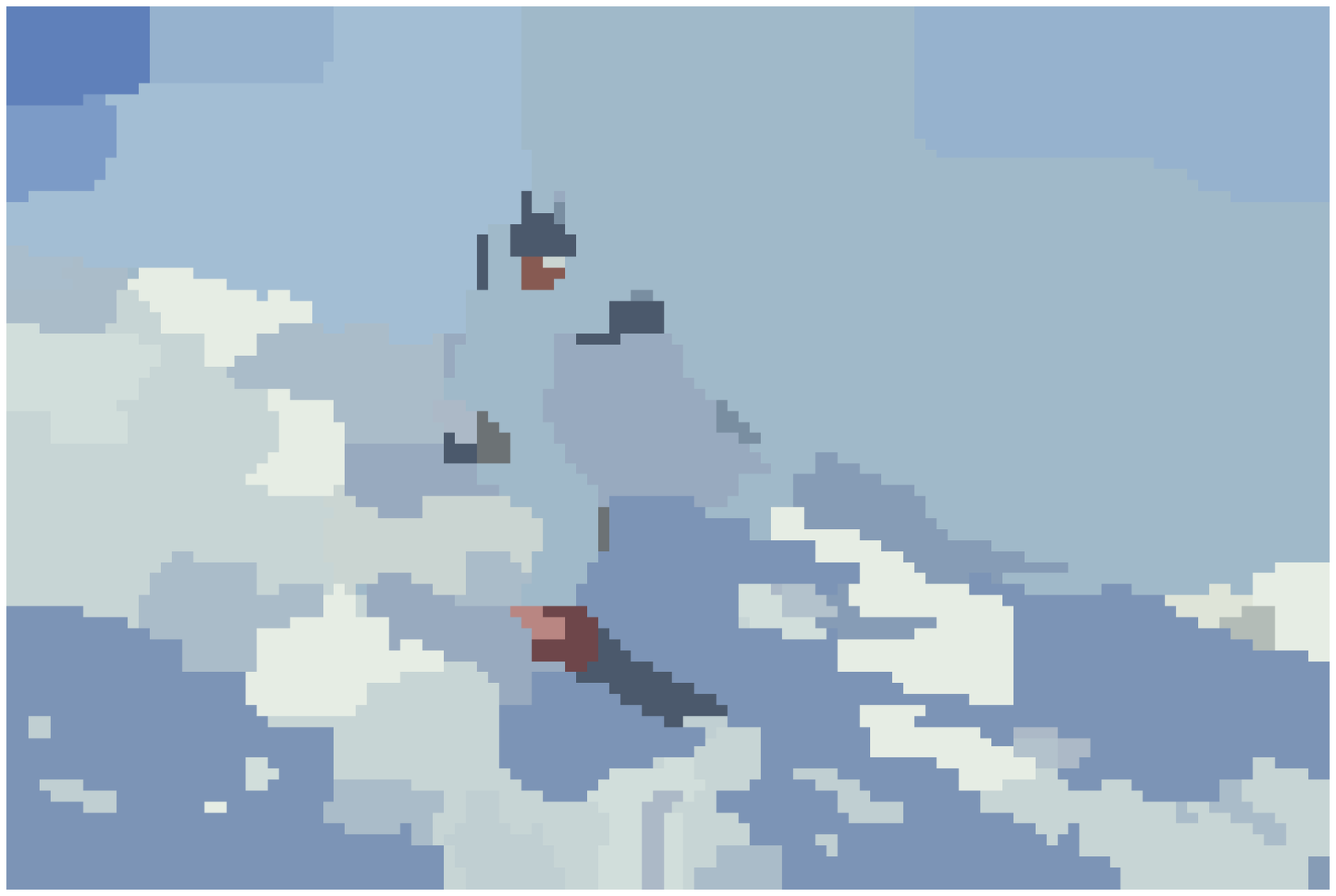} \\
				\includegraphics[width=1.1in]{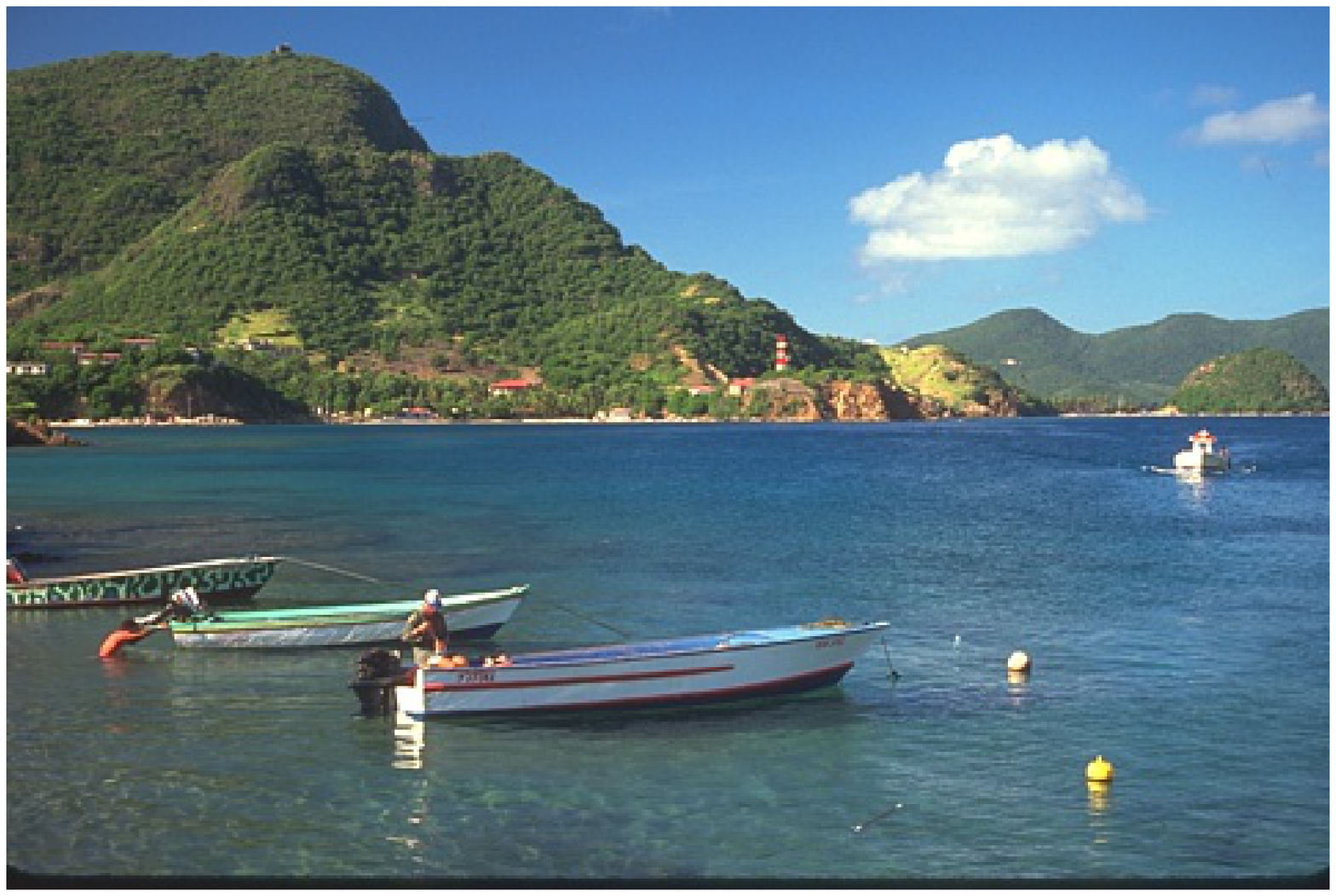} &
				\includegraphics[width=1.1in]{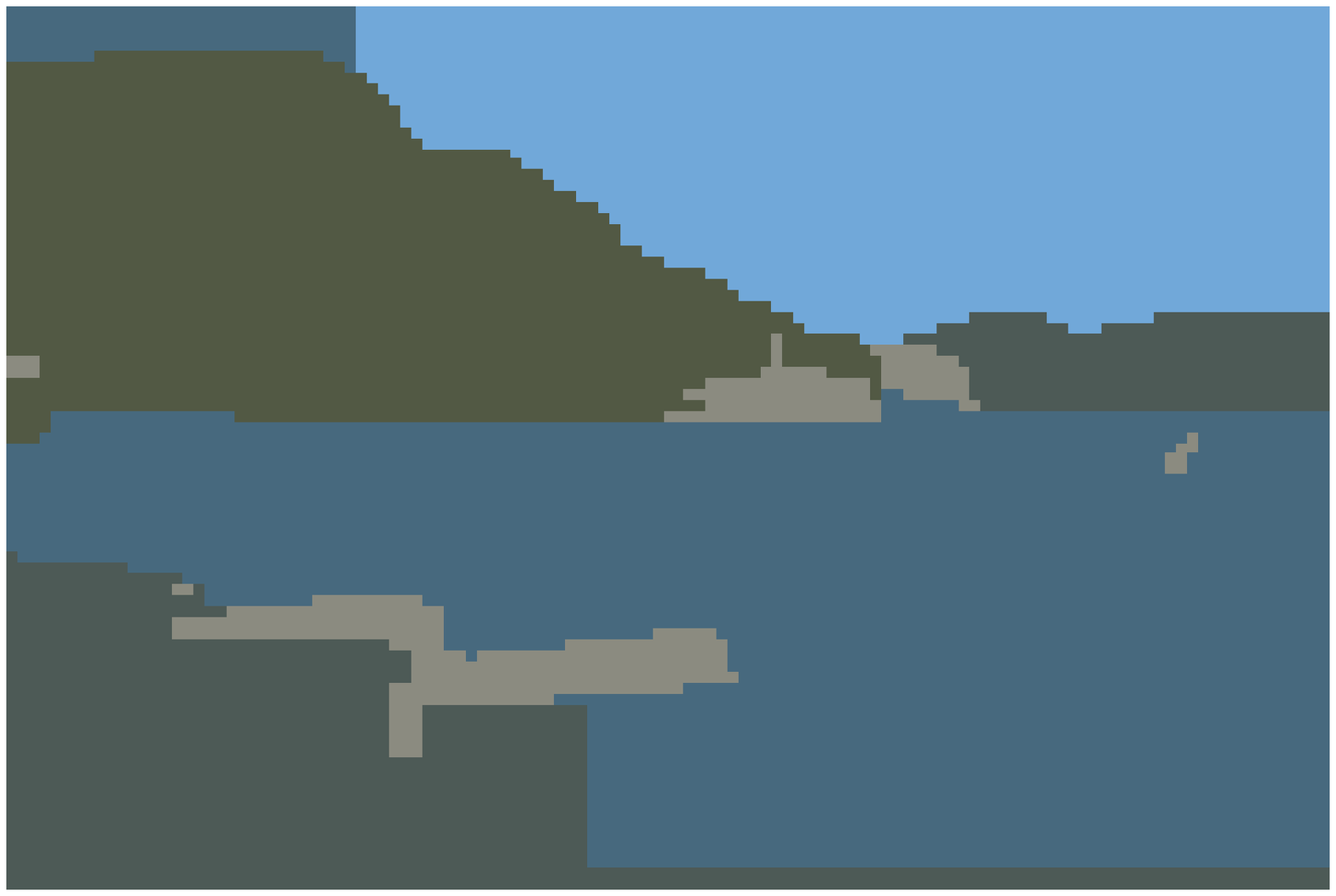} &
				\includegraphics[width=1.1in]{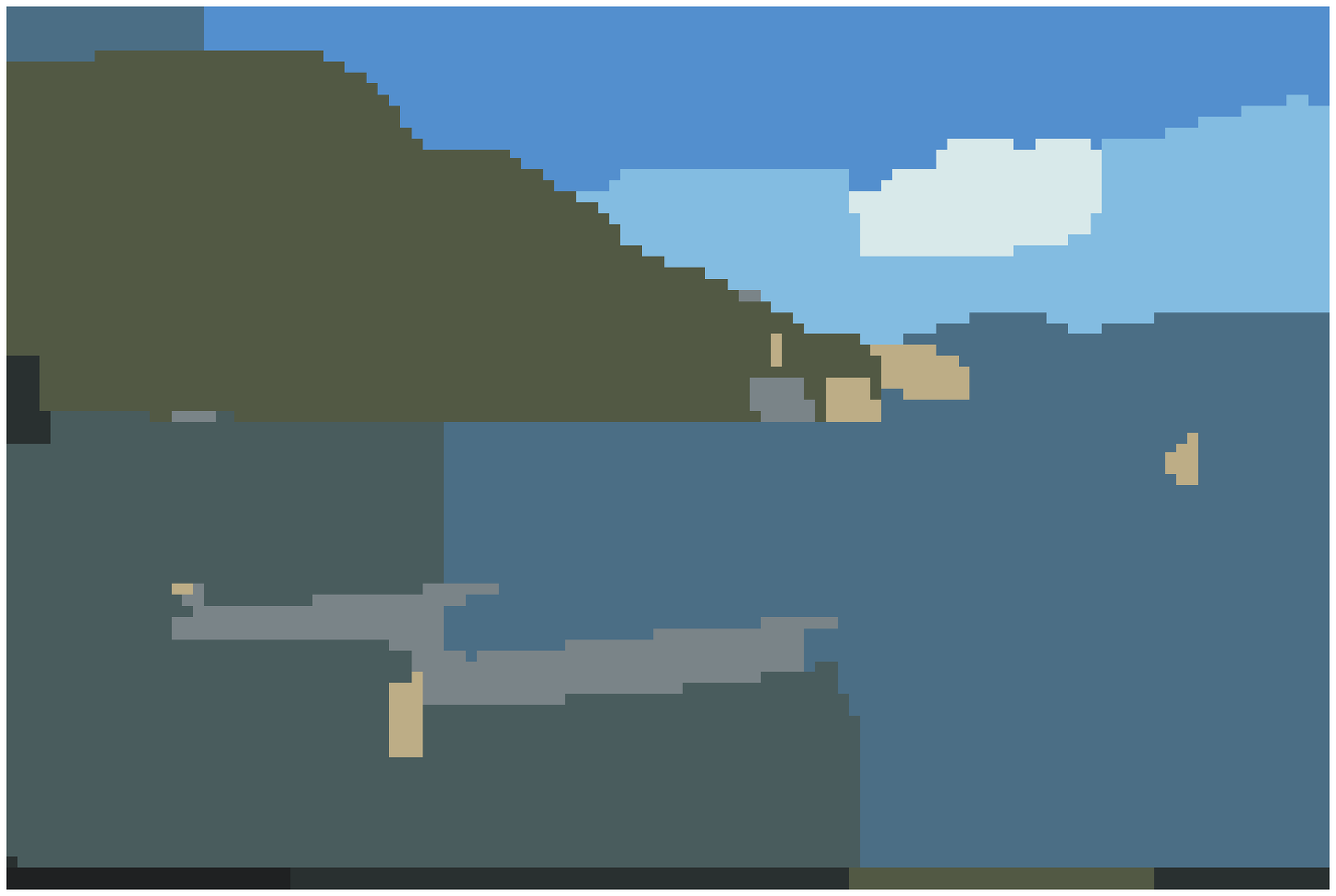} &
				\includegraphics[width=1.1in]{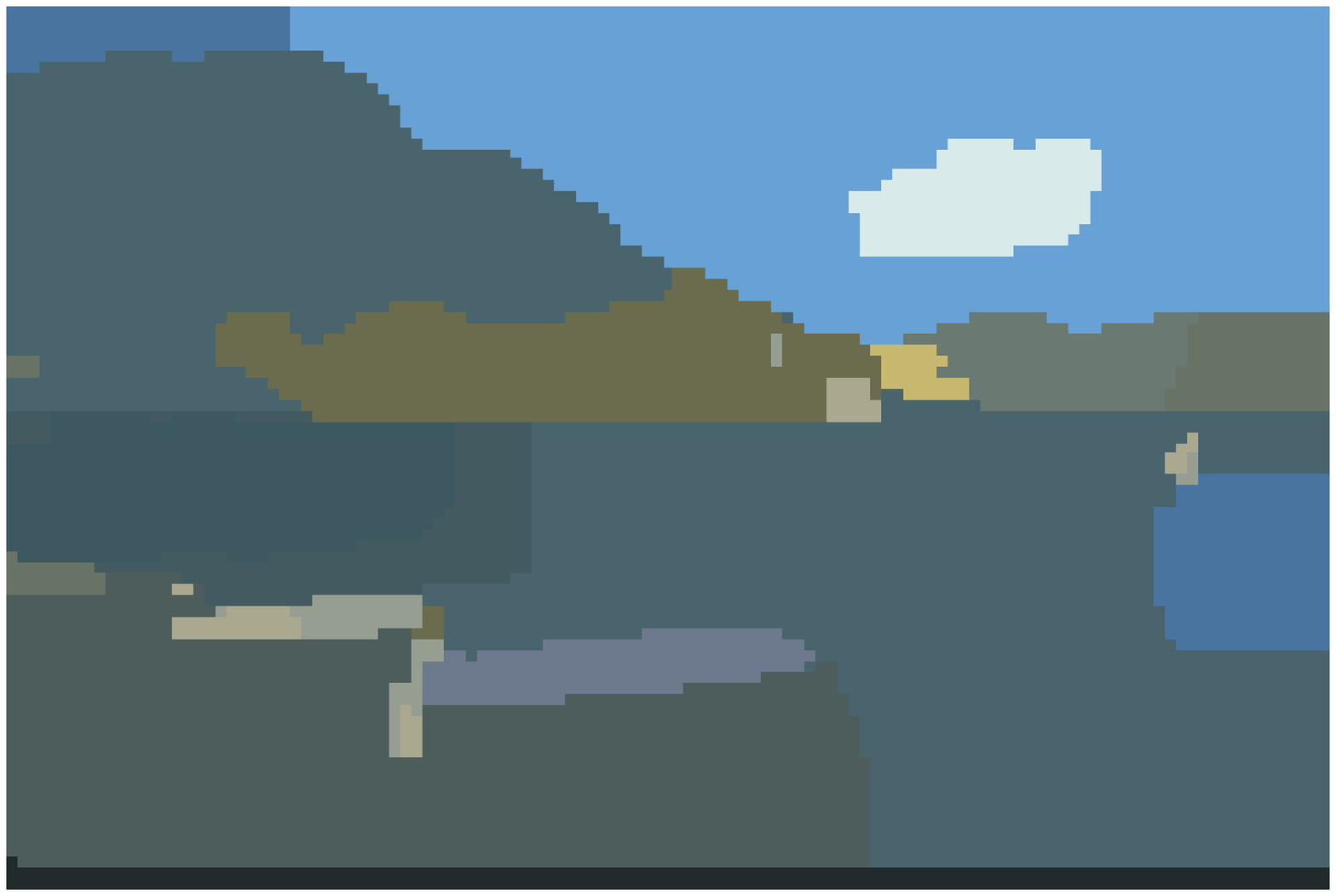} &
				\includegraphics[width=1.1in]{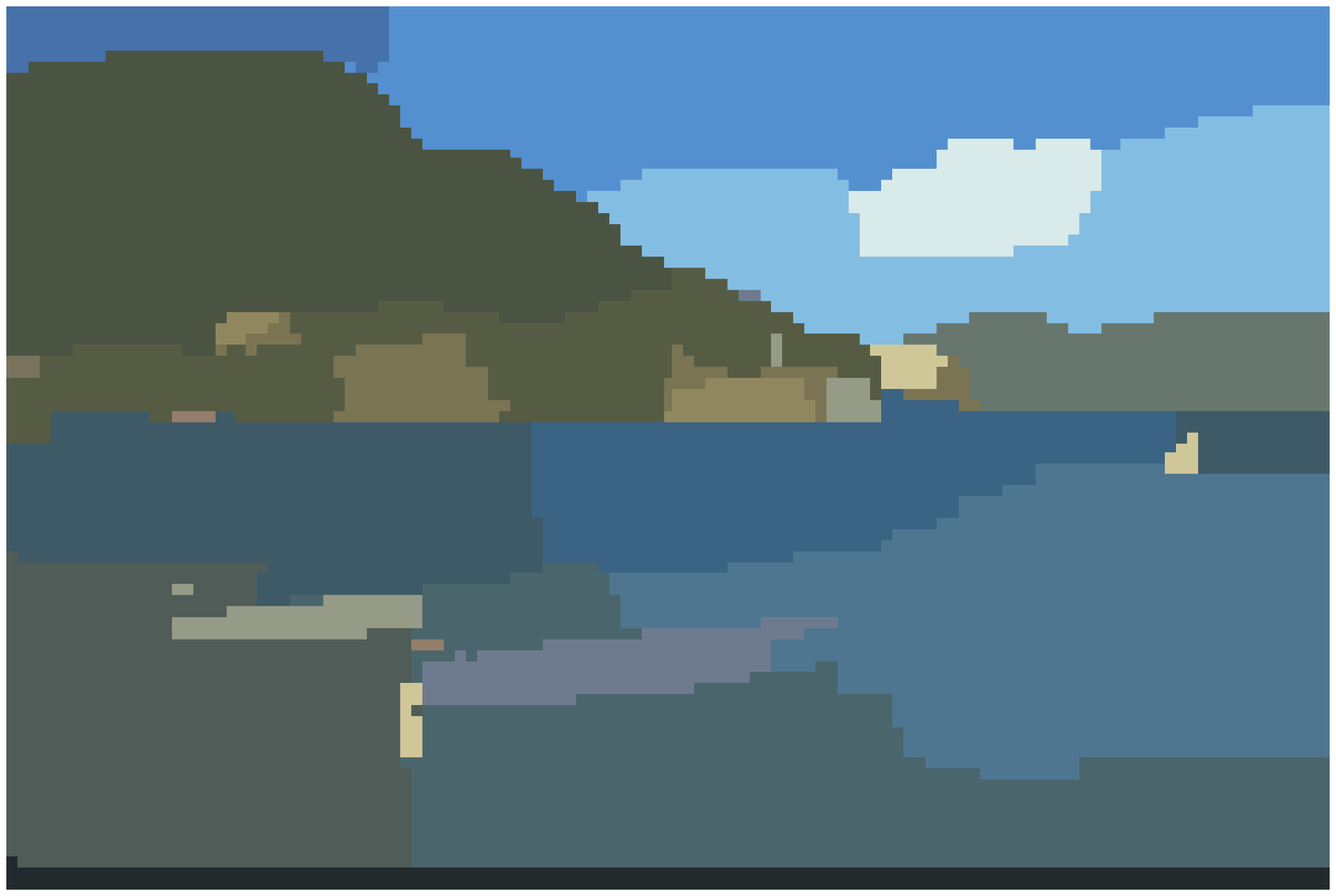} &
				\includegraphics[width=1.1in]{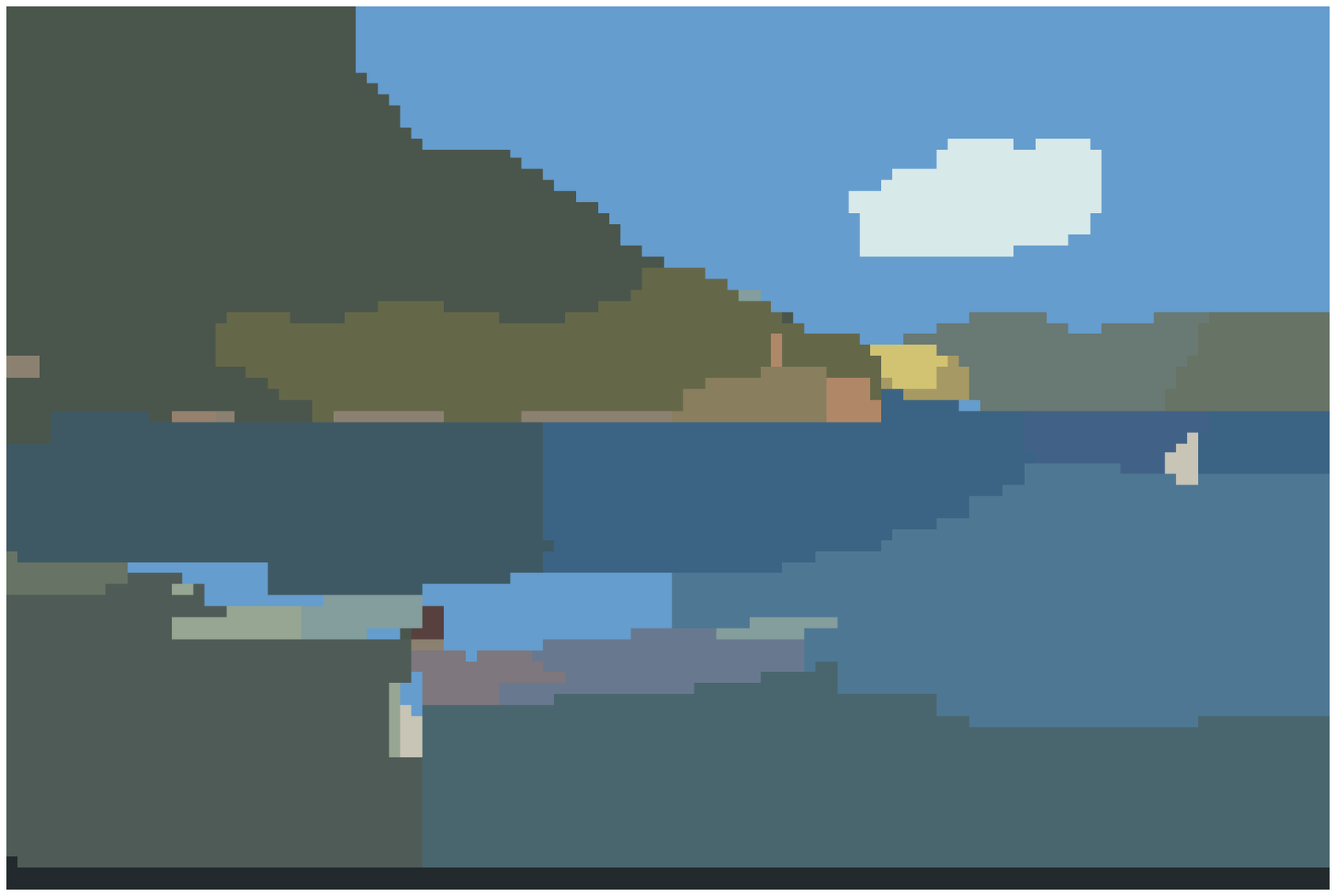} \\
				\includegraphics[width=1.1in]{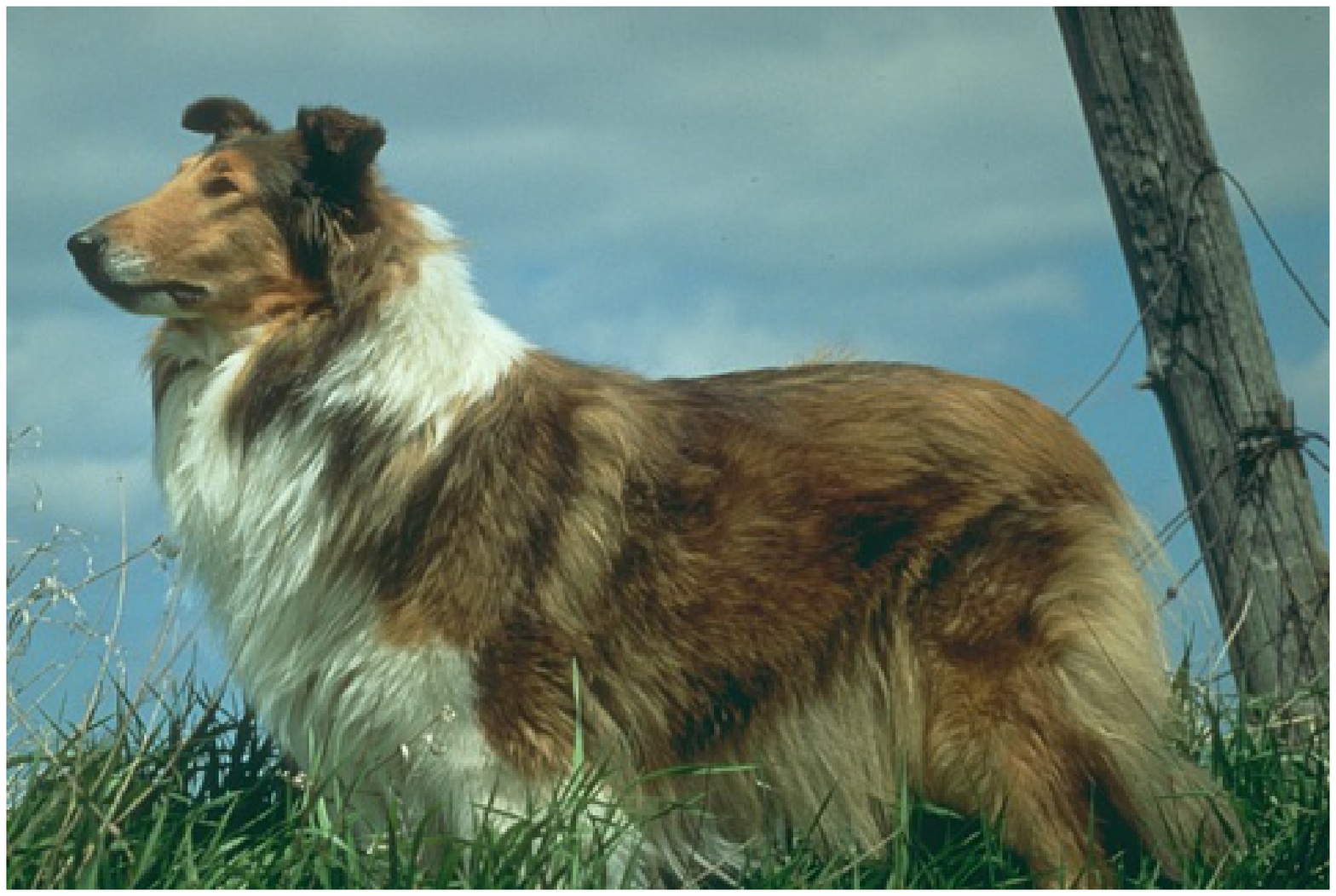} &
				\includegraphics[width=1.1in]{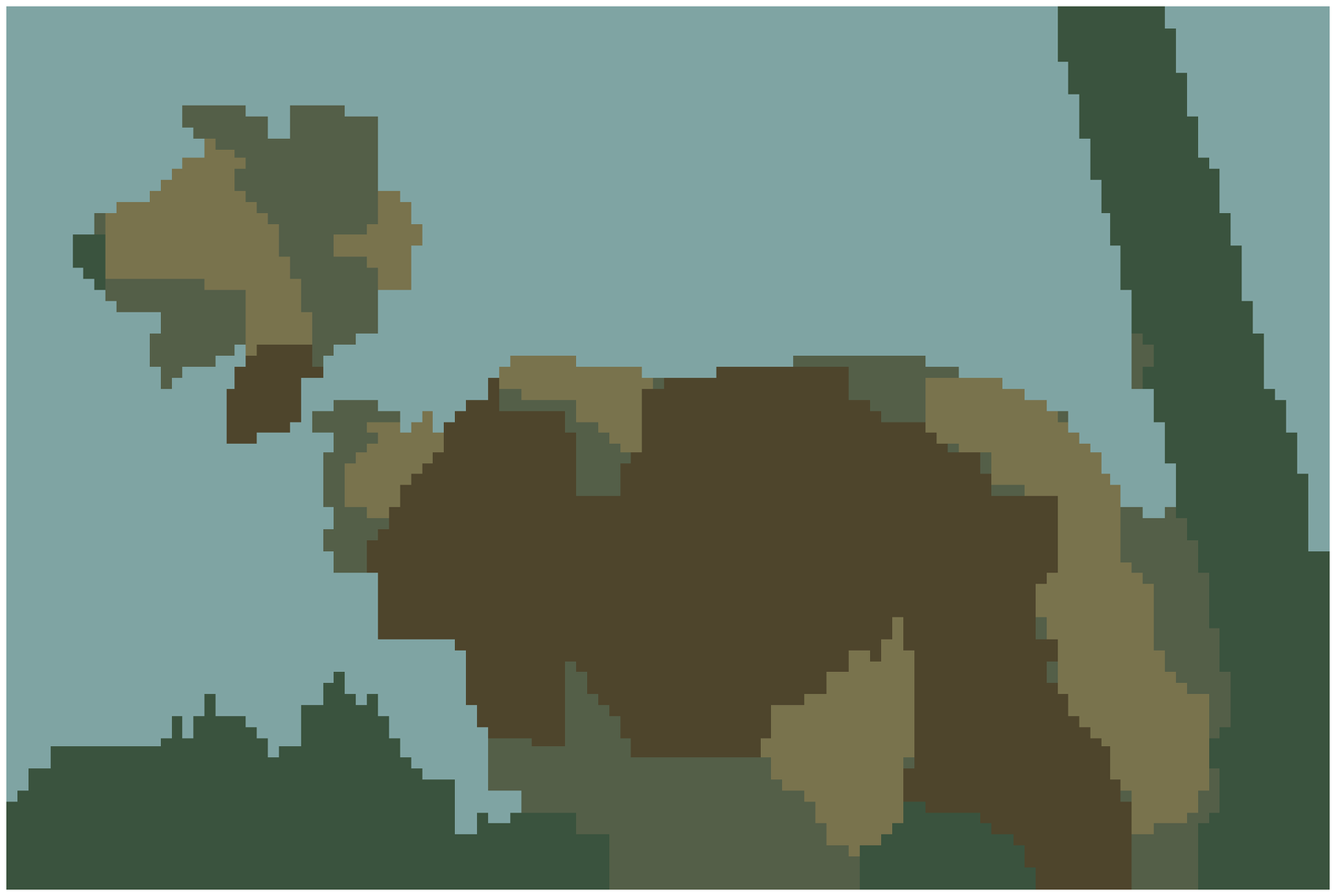} &
				\includegraphics[width=1.1in]{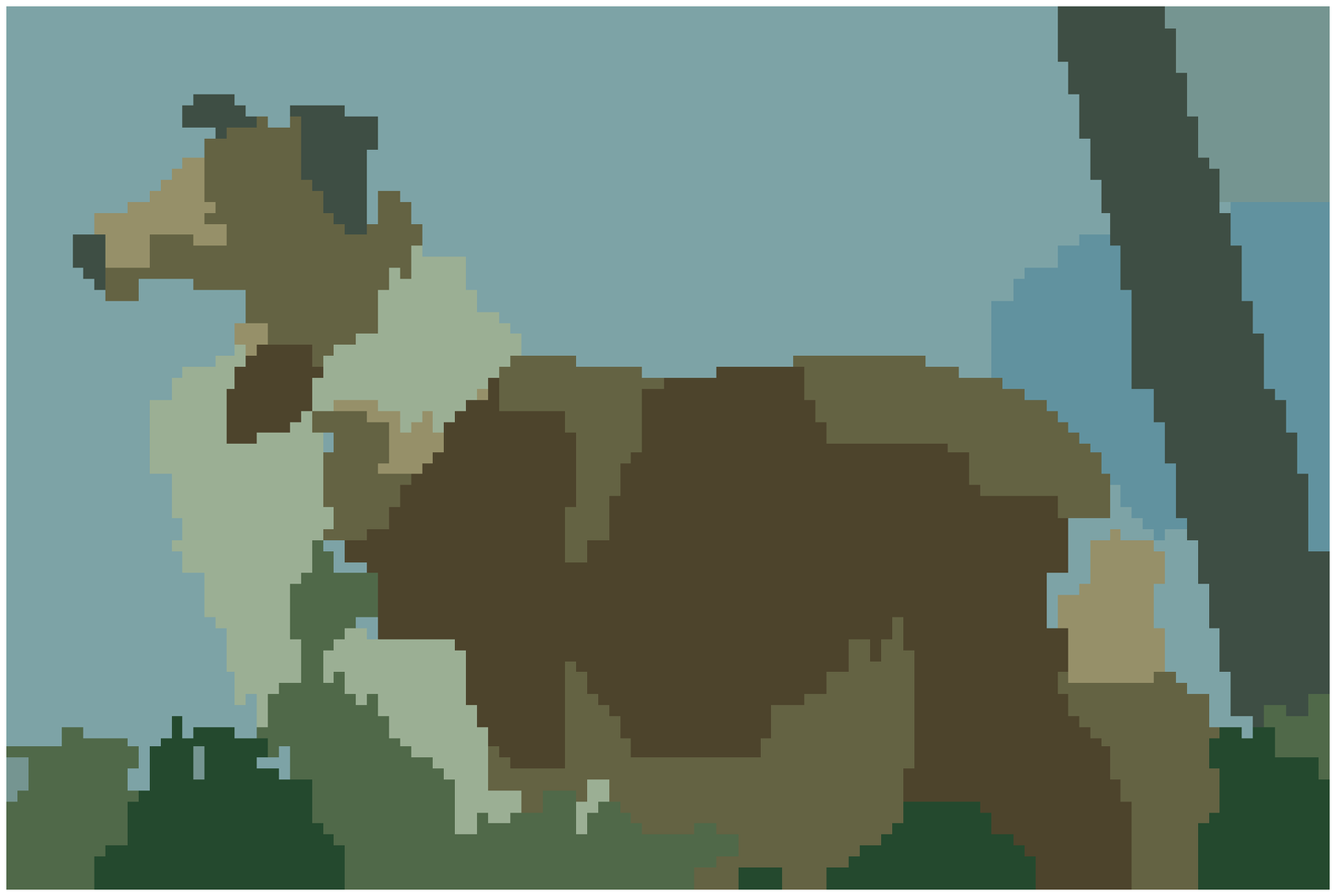} &
				\includegraphics[width=1.1in]{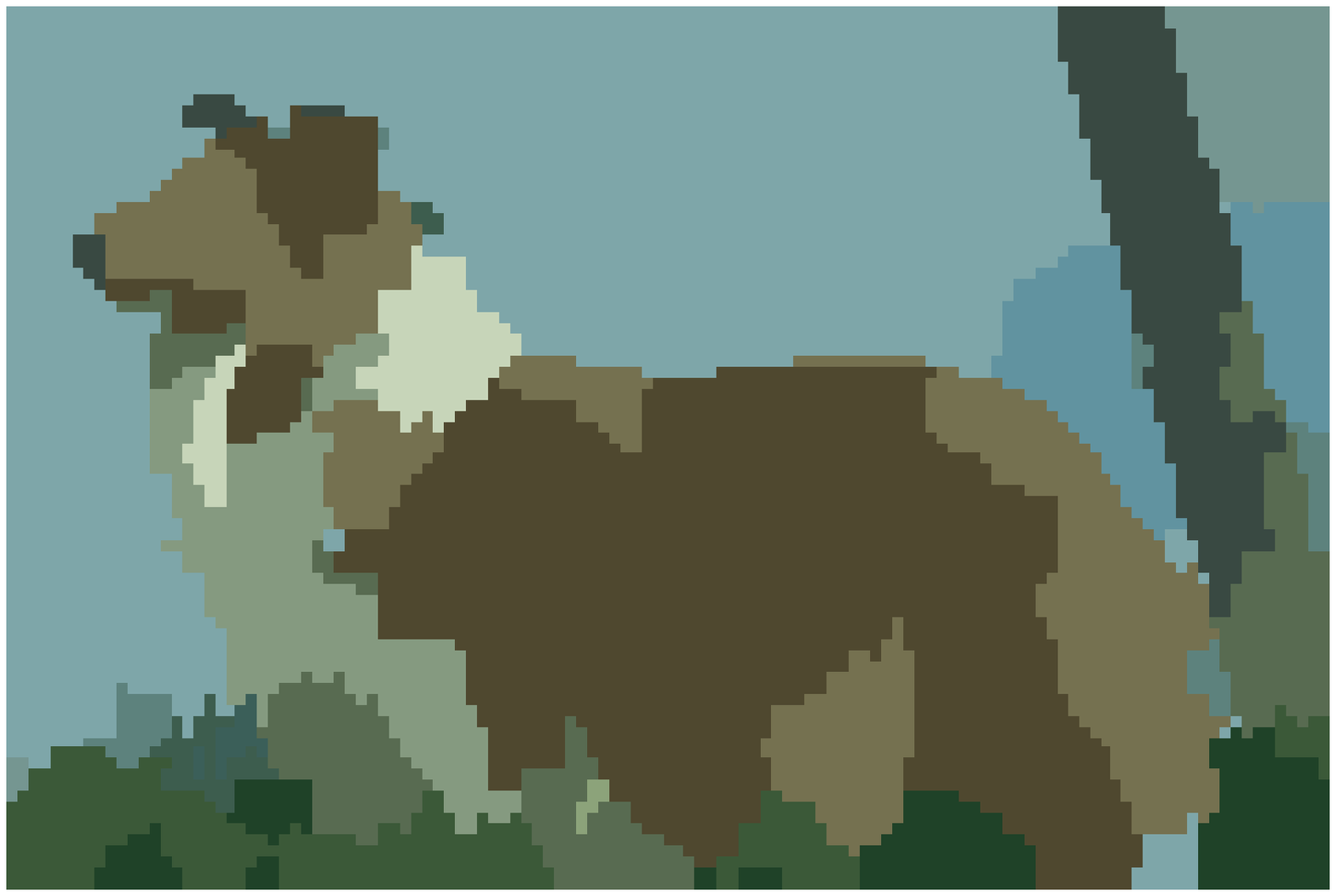} &
				\includegraphics[width=1.1in]{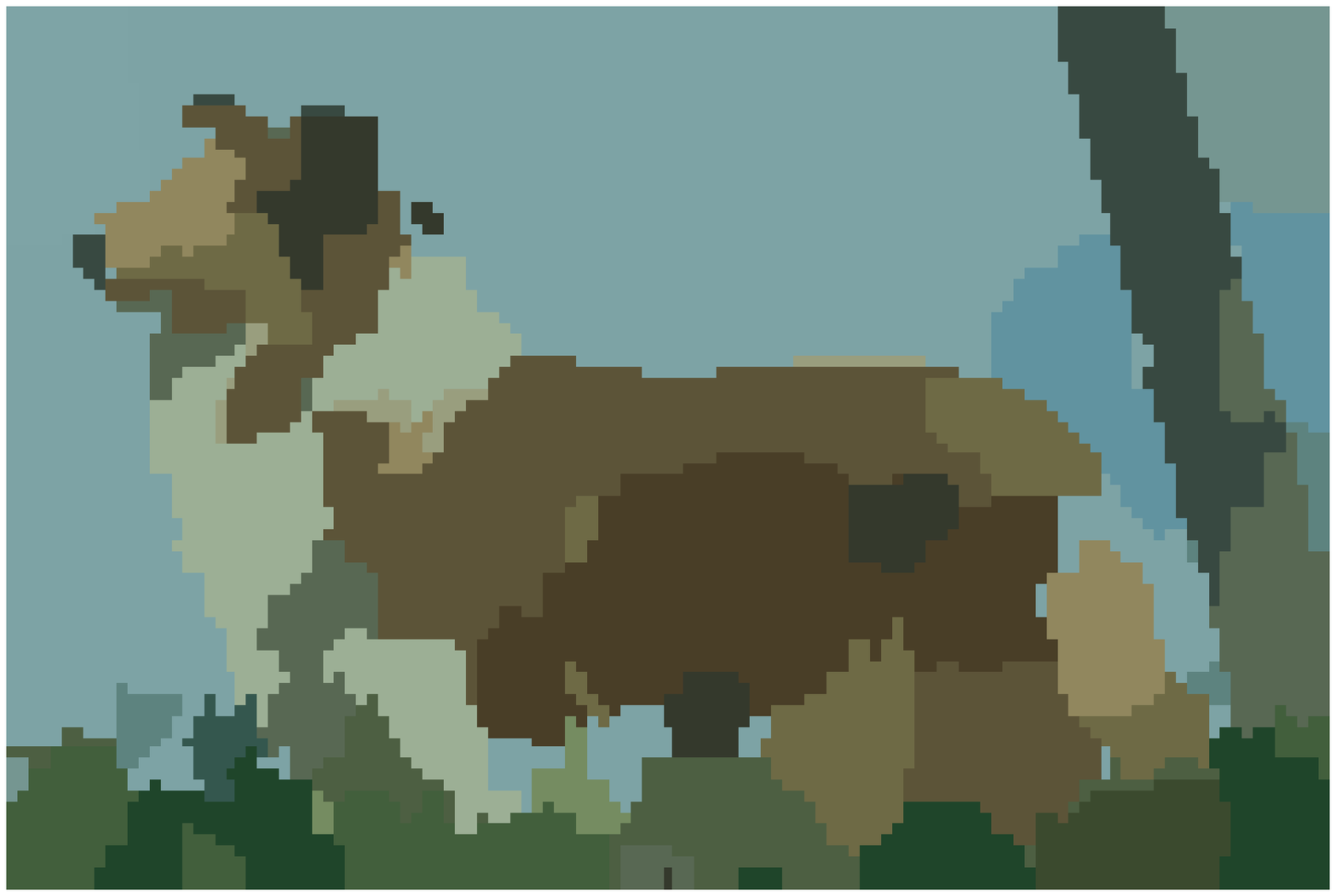} &
				\includegraphics[width=1.1in]{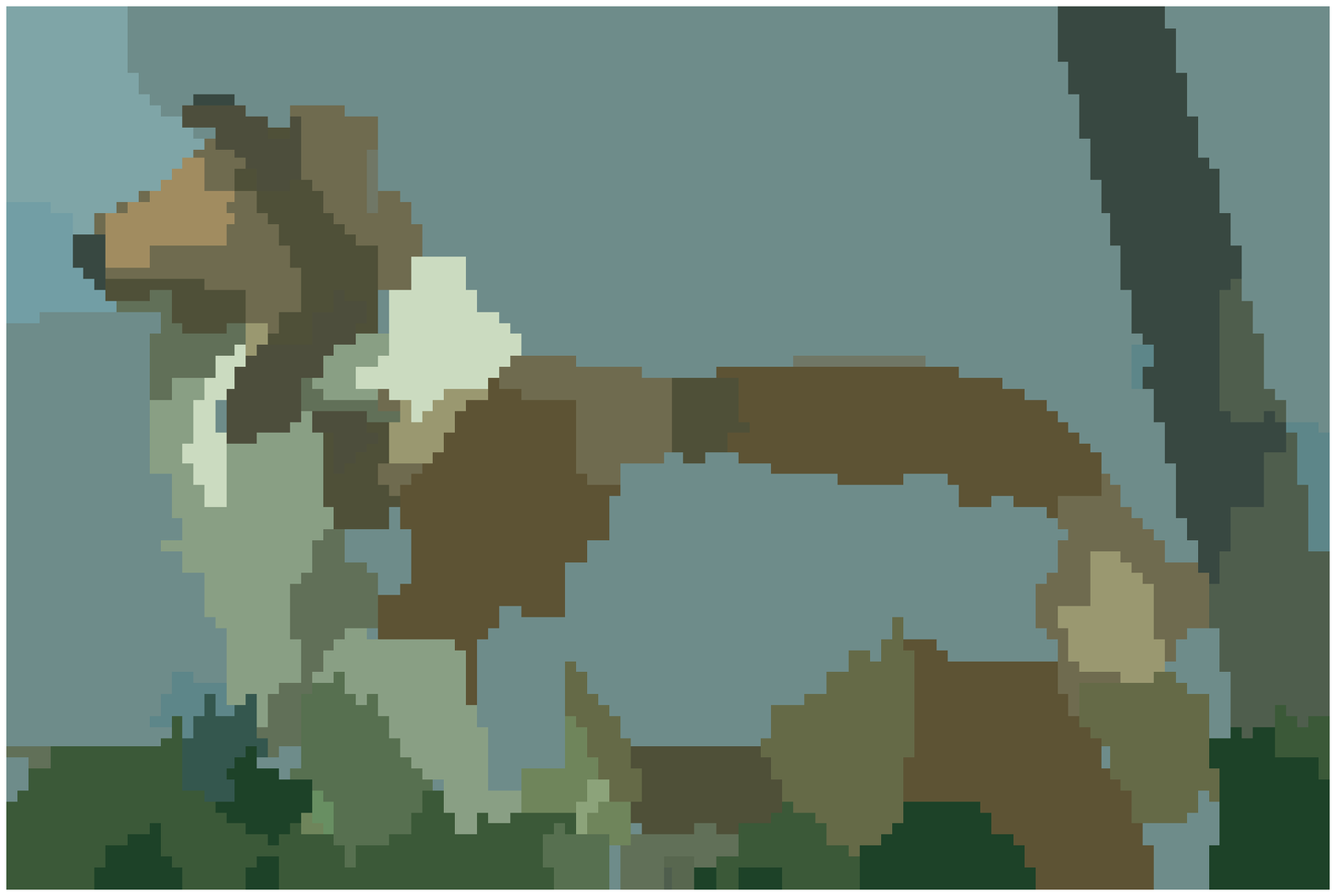} \\
				\includegraphics[width=1.1in]{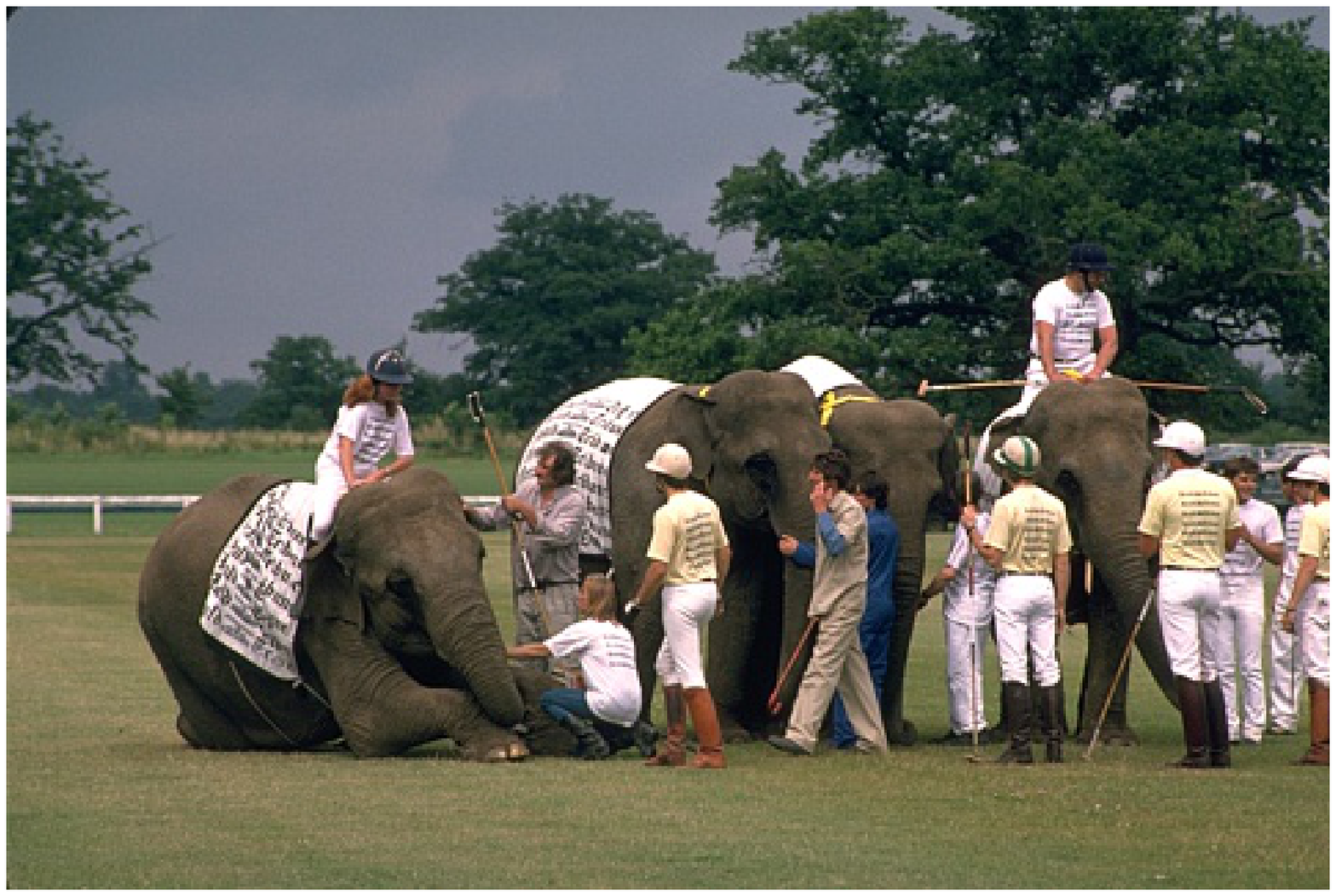} &
				\includegraphics[width=1.1in]{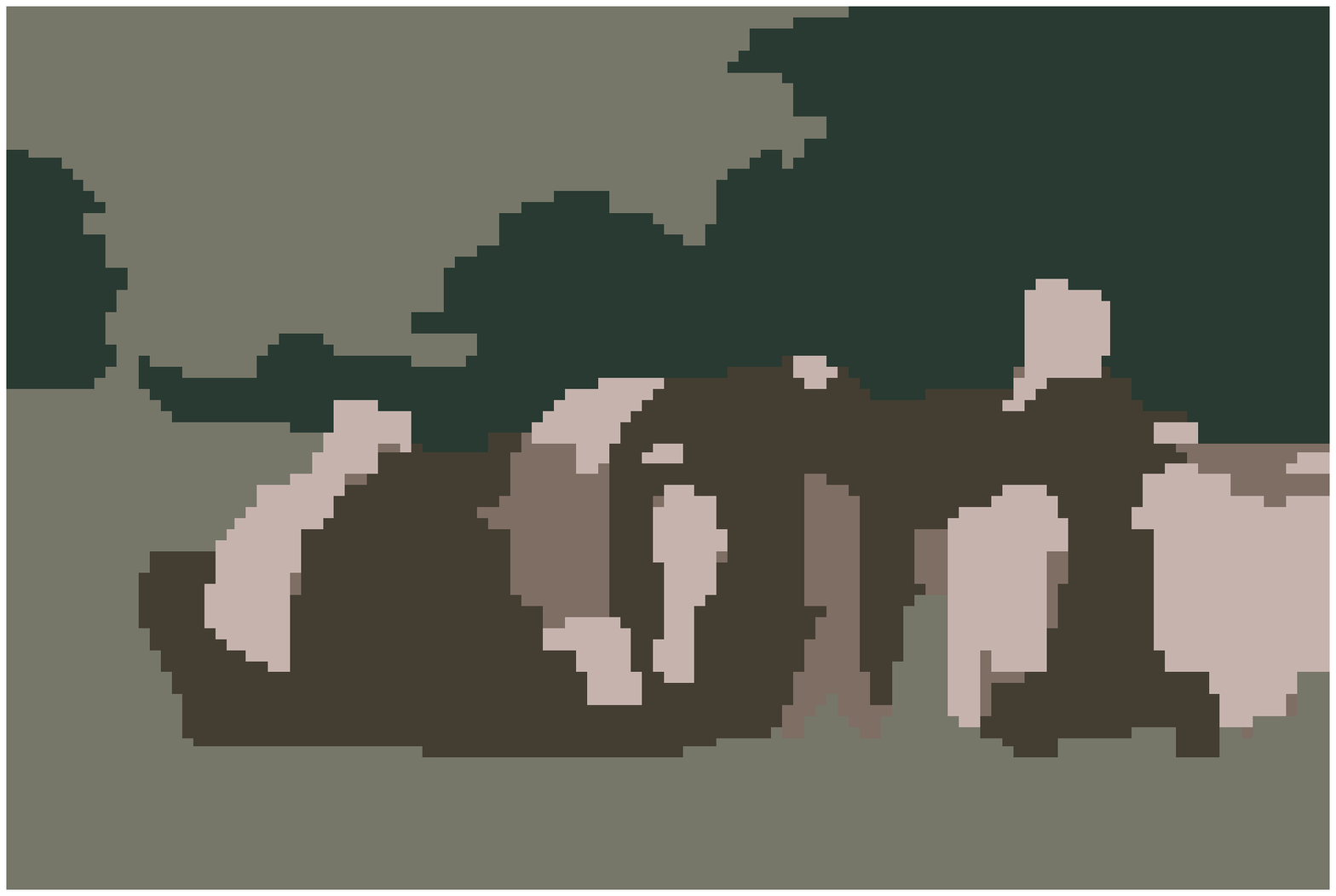} &
				\includegraphics[width=1.1in]{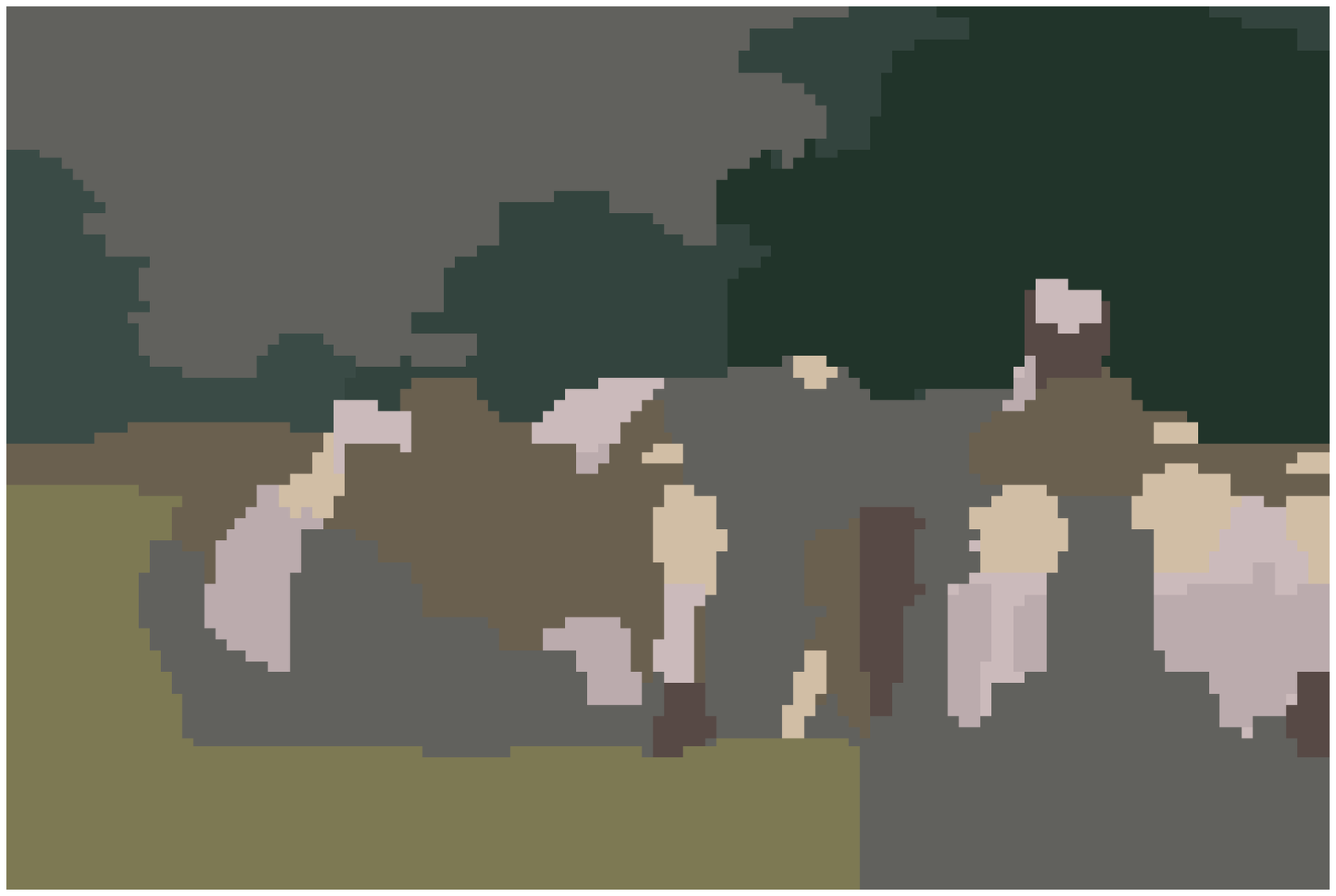} &
				\includegraphics[width=1.1in]{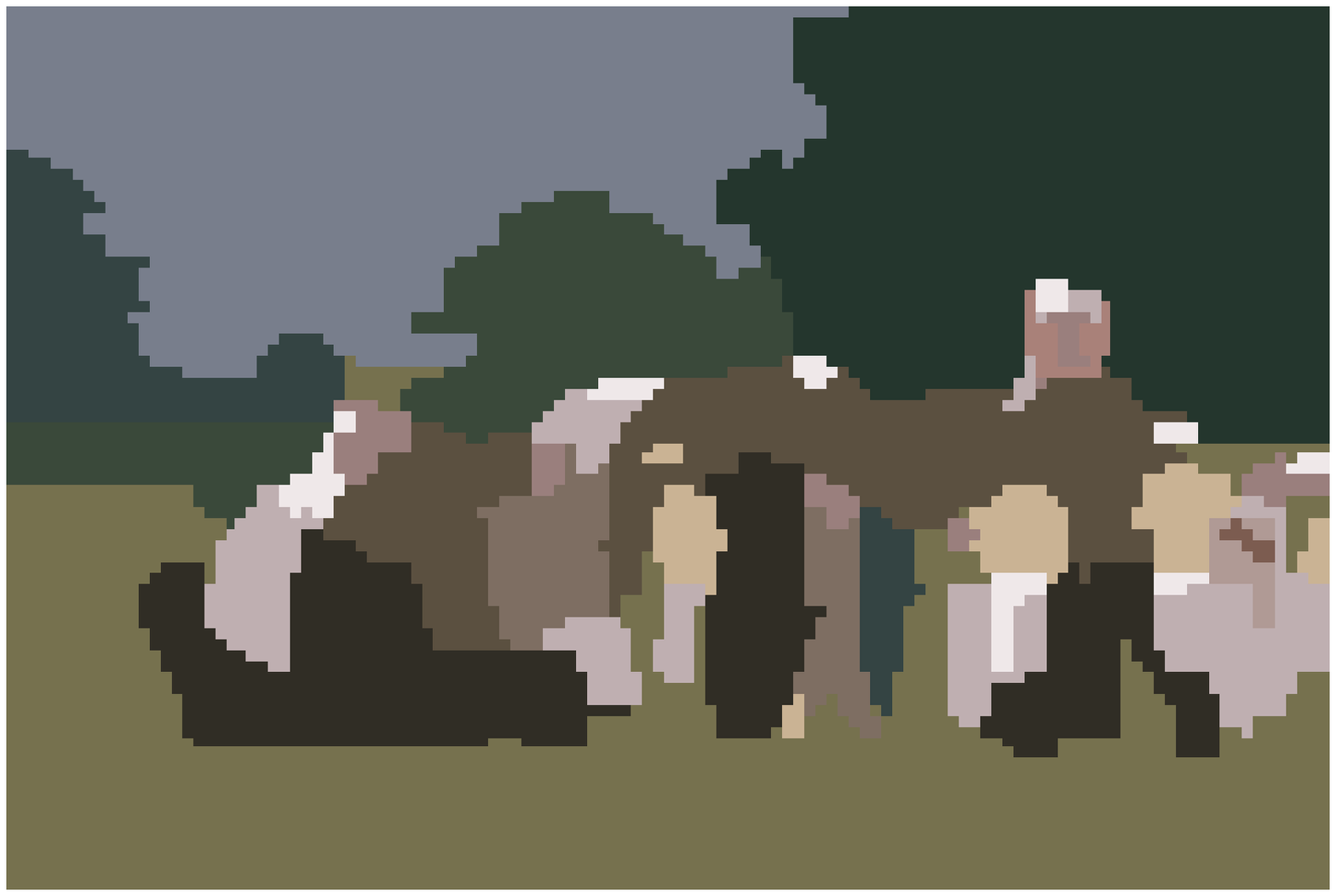} &
				\includegraphics[width=1.1in]{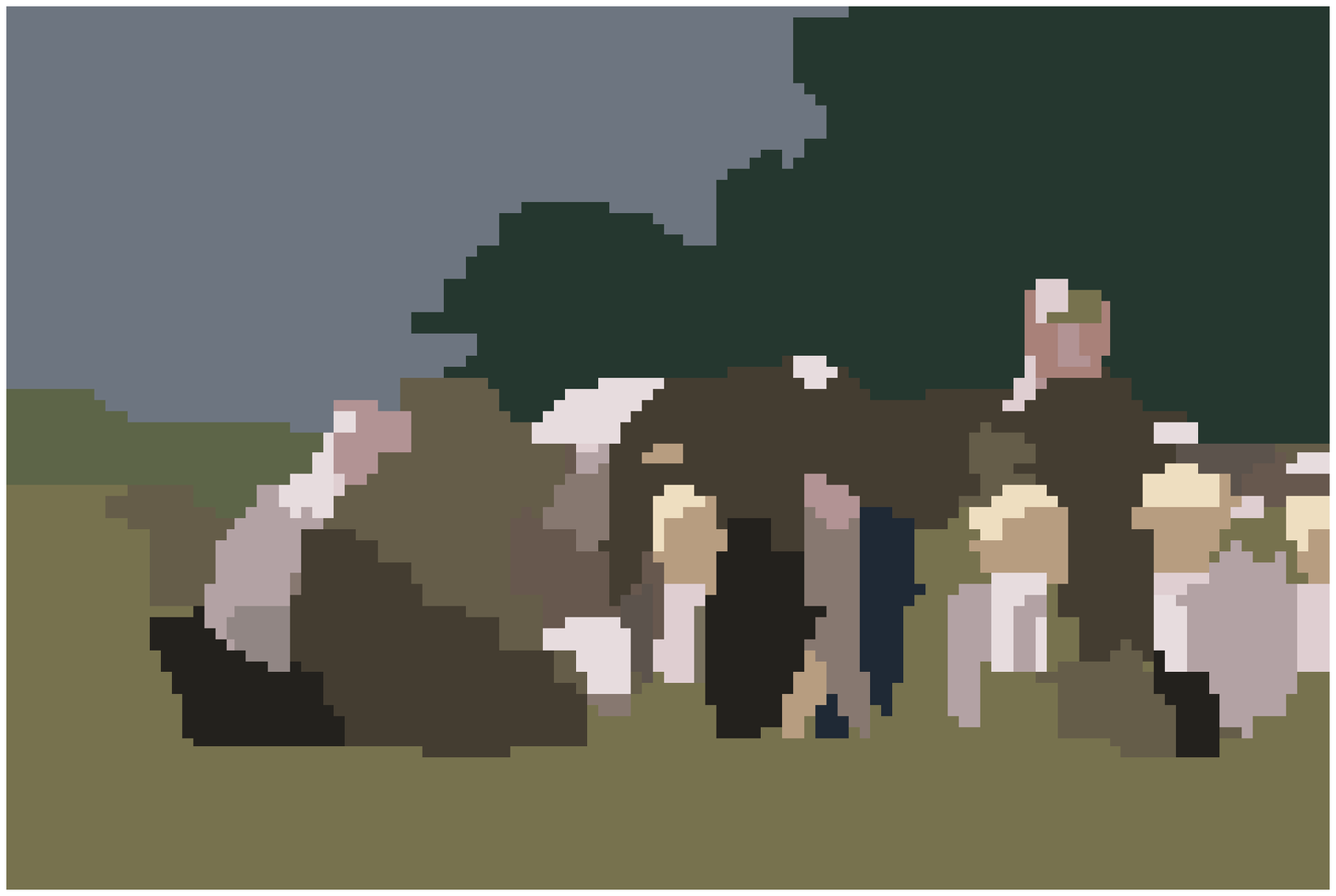} &
				\includegraphics[width=1.1in]{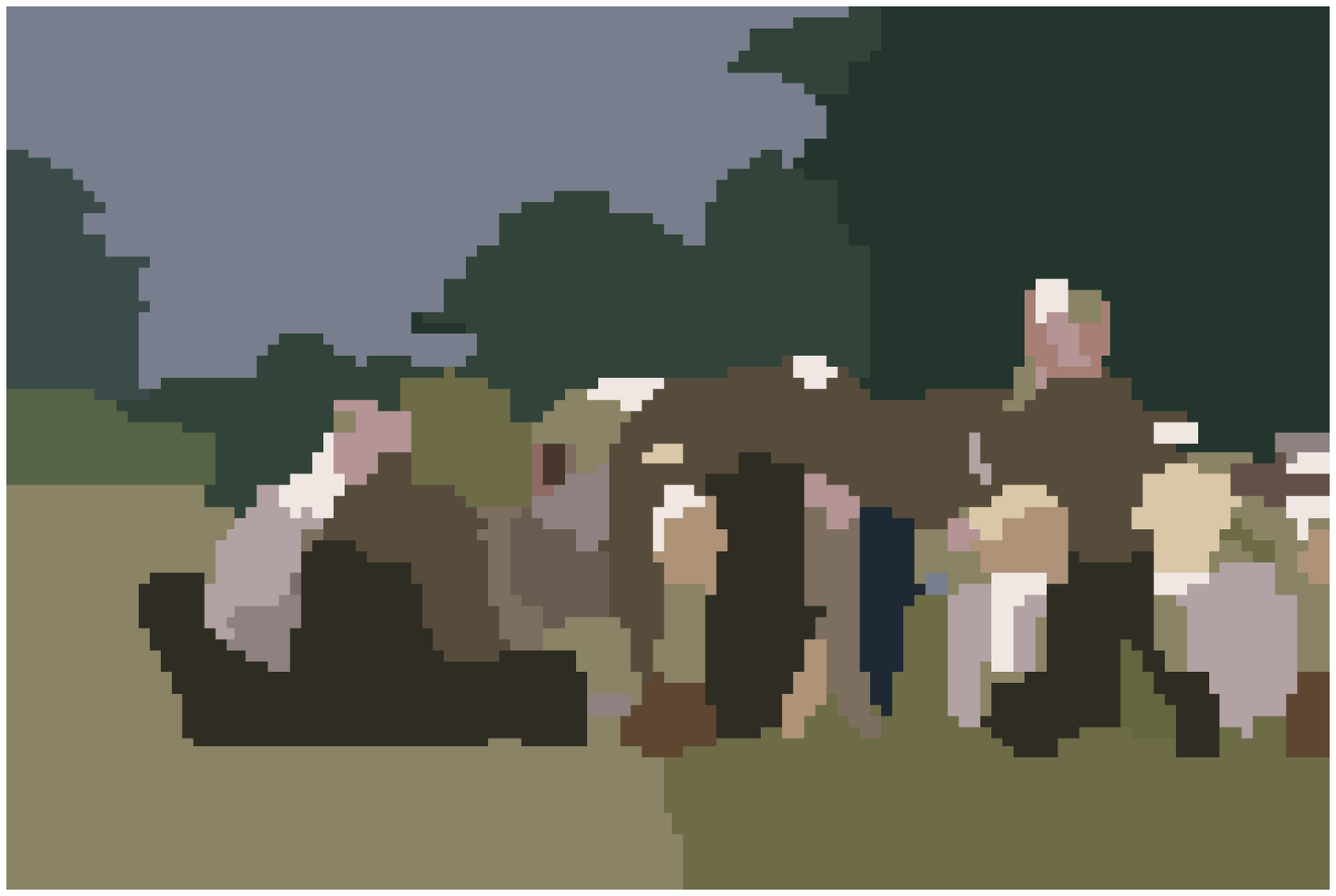} \\
			Original & $d=5$ & $d=10$ & $d=15$ & $d=20$ & $d=25$
		\end{tabular}
	\end{center}
		\vspace{-10pt}
		\caption{Selection of BSDS500 training images with PFE-based segmentations for various embedding dimensions $d$.}
\label{fig:BSDS500}
\end{figure*}

To construct a weighted graph for each image, we assign a graph vertex to each pixel (after downsampling images by a factor of $4$ in each dimension), and we only assign nonzero weights (according to the heat kernel applied to the RGB image data) to pixels that are in each other's $4$-neighborhood. To initialize $\mathbf{Y}$, we use the Gaussian Mixture Model (GMM) approach outlined in \cite{yu2015pfe}. We then follow Yu et al.'s \emph{dynamic} scheme that chooses the embedding dimension $d$ that produces the best performance out of dimensions in the range of $5$ to $25$. Figure \ref{fig:BSDS500} shows some of the BSDS500 training images along with examples of segmentation results via our implementation of PFE for various choices of embedding dimension. 

We use the three criteria outlined in the BSDS segmentation benchmark \cite{arbelaez2011cdh} to assess how similar each segmented image is to the ground truth segmentation that was manually created and is provided with the BSDS500 dataset. The criteria are: covering, which quantifies overlap between segmentations; probabilistic rand index (PRI), which quantifies the "compatibility" of segmentations; and variation of information (VI), which describes the average conditional entropy of two segmentations. Larger covering and PRI values indicate better performance, whereas smaller VI values indicate better performance. The results are shown in Table \ref{tab:results}, along with the results of Yu et al.'s PFE implementation and three other methods tested in \cite{yu2015pfe}: normalized cuts (NCut), spectral clustering (SC), and weighted spectral clustering (WSC). As we can see, our efficient implementation of PFE yields similar covering and PRI values to the results reported in \cite{yu2015pfe}, but slightly worse VI values. However, the standard deviation of our VI performance measure was computed to be $0.67$. If Yu et al. had a similar standard deviation in VI values (which was not reported in \cite{yu2015pfe}), it is likely that the difference in VI values between our implementation and Yu's is not statistically significant.

\begin{figure}[ht]
\centering
\includegraphics[width=3in]{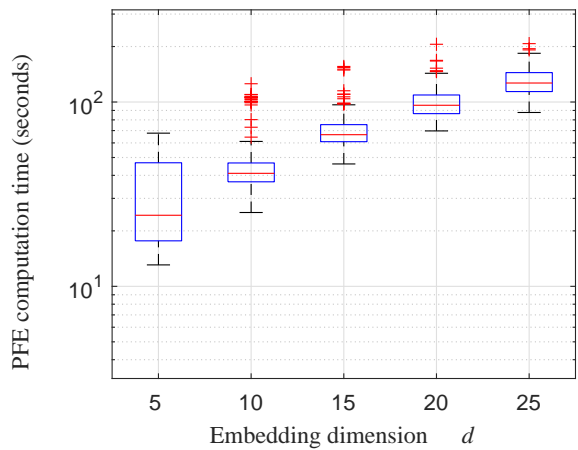}
\caption{Box plots of computation times required to efficiently compute PFE on BSDS500 images. Red $+$'s indicate outliers.}
\label{fig:timing}
\end{figure}

Finally, in Figure \ref{fig:timing}, we show box-and-whisker plots of the time (in seconds) required to compute PFE for the 200 BSDS500 training images, for a range of embedding dimensions. All computations are done in MATLAB. As can be seen in Figure \ref{fig:timing}, our implementation typically requires anywhere from $0.5$--$2$ minutes, with more time required for higher embedding dimension. Yu et al. \cite{yu2015pfe} report an average computing time of 15 minutes per image; however, as of the time of this article, they have not released any code, making it impossible for us to do a direct comparison. 

\begin{table}[h] 
\begin{center}
\begin{tabular}{|c|c|c|c|}
    \hline
    Method & Covering & PRI & VI \\
    \hline\hline
		NCut & 0.40 & 0.76 & 2.39 \\
		\hline
		SC & 0.44 & 0.77 & 2.24 \\
		\hline
		WSC & 0.44 & 0.77 & 2.21 \\
		\hline
		Yu-PFE & 0.52 & $\textbf{0.79}$ & $\textbf{1.91}$ \\
		\hline\hline
		Ours & $\textbf{0.53}$ & $\textbf{0.79}$ & 2.10 \\
		\hline
\end{tabular}
\end{center}
\caption{Comparison of Normalized Cut (NCut), spectral clustering (SC), weighted spectral clustering (WSC), Yu et al. implementation of PFE (Yu-PFE), and our efficient PFE method (Ours) on BSDS5000. Best results for each performance measure highlighted in bold.}
\label{tab:results}
\end{table}

\section{Conclusion}
Piecewise Flat Embeddings, originally proposed in \cite{yu2015pfe}, provide powerful data representations that can be used for clustering and image segmentation. However, the description of the original algorithm can be improved to incorporate two efficiencies: the reduction of the Split-Bregman iteration to work on more compact matrices, and the use of a preconditioned conjugate gradient solver to rapidly solve the linear least squares problem at the heart of each inner loop. We showed that with these efficiencies, we can replicate the image segmentation experiment performed in \cite{yu2015pfe} in a manner that yields similar performance measures but only requires $0.5$--$2$ minutes per image instead of the $15$ minutes previously reported. 

\section*{Code}
A prototype implementation of our algorithm for efficiently computing PFE is available at MATLAB Central (\url{http://www.mathworks.com/matlabcentral/}) under File ID \#59763.

\ifCLASSOPTIONcaptionsoff
  \newpage
\fi

\bibliographystyle{IEEEtran}
\bibliography{refs}

\begin{thebibliography}{1}
\providecommand{\url}[1]{#1}
\csname url@samestyle\endcsname
\providecommand{\newblock}{\relax}
\providecommand{\bibinfo}[2]{#2}
\providecommand{\BIBentrySTDinterwordspacing}{\spaceskip=0pt\relax}
\providecommand{\BIBentryALTinterwordstretchfactor}{4}
\providecommand{\BIBentryALTinterwordspacing}{\spaceskip=\fontdimen2\font plus
\BIBentryALTinterwordstretchfactor\fontdimen3\font minus
  \fontdimen4\font\relax}
\providecommand{\BIBforeignlanguage}[2]{{%
\expandafter\ifx\csname l@#1\endcsname\relax
\typeout{** WARNING: IEEEtran.bst: No hyphenation pattern has been}%
\typeout{** loaded for the language `#1'. Using the pattern for}%
\typeout{** the default language instead.}%
\else
\language=\csname l@#1\endcsname
\fi
#2}}
\providecommand{\BIBdecl}{\relax}
\BIBdecl

\bibitem{yu2015pfe}
Y.~Yu, C.~Fang, and Z.~Liao, ``Piecewise flat embedding for image
  segmentation,'' in \emph{Proceedings of the IEEE International Conference on
  Computer Vision}, 2015, pp. 1368--1376.

\bibitem{belkin2003led}
M.~Belkin and P.~Niyogi, ``Laplacian eigenmaps for dimensionality reduction and
  data representation,'' \emph{Neural computation}, vol.~15, no.~6, pp.
  1373--1396, 2003.

\bibitem{goldstein2009sbm}
T.~Goldstein and S.~Osher, ``The split {B}regman method for {L}1-regularized
  problems,'' \emph{SIAM journal on imaging sciences}, vol.~2, no.~2, pp.
  323--343, 2009.

\bibitem{lai2014smo}
R.~Lai and S.~Osher, ``A splitting method for orthogonality constrained
  problems,'' \emph{Journal of Scientific Computing}, vol.~58, no.~2, pp.
  431--449, 2014.

\bibitem{barrett1994tsl}
R.~Barrett, M.~W. Berry, T.~F. Chan, J.~Demmel, J.~Donato, J.~Dongarra,
  V.~Eijkhout, R.~Pozo, C.~Romine, and H.~Van~der Vorst, \emph{Templates for
  the Solution of Linear Systems: Building Blocks for Iterative Methods}.\hskip
  1em plus 0.5em minus 0.4em\relax {SIAM}, 1994, vol.~43.

\bibitem{arbelaez2011cdh}
P.~Arbelaez, M.~Maire, C.~Fowlkes, and J.~Malik, ``Contour detection and
  hierarchical image segmentation,'' \emph{IEEE Transactions on Pattern
  Analysis and Machine Intelligence}, vol.~33, no.~5, pp. 898--916, 2011.

\end{thebibliography}

\end{document}